\documentclass{article}

\PassOptionsToPackage{table}{xcolor}
\usepackage{microtype}
\usepackage{graphicx}
\usepackage{booktabs}
\usepackage{hyperref}

\usepackage[preprint]{icml2027}
\hypersetup{pdfsubject={}}
\makeatletter
\renewcommand{\ICML@preprint}{\textit{Preprint.}}
\makeatother
\usepackage{amsmath}
\usepackage{amssymb}
\usepackage{array}
\usepackage{tikz}
\usetikzlibrary{arrows.meta,calc,positioning}

\newcommand{\mf}{\mathrm{mf}}
\definecolor{ttaccent}{RGB}{31,95,115}
\definecolor{ttcut}{RGB}{155,47,47}
\definecolor{ttkeep}{RGB}{44,110,63}
\definecolor{ttpart}{RGB}{138,90,0}
\newcommand{\cellLost}[1]{\cellcolor{ttaccent!14}#1}
\newcommand{\cellPart}[1]{\cellcolor{ttpart!18}#1}
\newcommand{\cellNA}[1]{\cellcolor{black!5}\textcolor{black!60}{#1}}
\newcolumntype{L}[1]{>{\raggedright\arraybackslash}p{#1}}
\icmltitlerunning{Token Tells: Prompt Tokens That Copy a Steering Vector Need Its Direction}

\begin{document}

\twocolumn[
  \icmltitle{Token Tells: Prompt Tokens That Copy a Steering Vector \\ Need Its Direction}

  \icmlsetsymbol{equal}{*}

  \begin{icmlauthorlist}
    \icmlauthor{Aniket Ghosh}{equal,neu}
    \icmlauthor{Mohamed Amine Merzouk}{equal,mila,mcgill}
    \icmlauthor{Adam Oberman}{mila,mcgill}
  \end{icmlauthorlist}
  \icmlaffiliation{neu}{Northeastern University}
  \icmlaffiliation{mila}{Mila -- Quebec AI Institute}
  \icmlaffiliation{mcgill}{McGill University}
  \icmlcorrespondingauthor{Aniket Ghosh}{ghosh.anik@northeastern.edu}

  \icmlkeywords{interpretability, steering vectors, refusal, prompt optimization, backdoors}

  \vskip 0.3in
]

\printAffiliationsAndNotice{\icmlEqualContribution}

\begin{abstract}
We introduce \emph{token steering}, a search for a few prompt tokens (a \emph{slot}) that make an unmodified language model behave like a modified one, and a \emph{route test}, which removes the modification's direction from the model and rescores the slot.
On gemma-2b-it, Llama-3-8B-Instruct and Yi-6B-Chat, three searched tokens make the model refuse harmless requests, and removing the refusal direction takes all $62$ slots we judged to $0.004$ or less on a harm-grounded judge, while removing a random direction does not.
Plain instructions depend on the direction only partly: on Llama-3-8B-Instruct, instructions that state a harm reason keep most of their harm-grounded refusal without it.
Slots that imitate a Spanish SAE feature, a Spanish steering vector, function vectors, a French backdoor and an evil persona vector also need their direction, and for all but the persona plain instructions need it too.
Refusal moves between models through text alone: slots searched against a steered model's saved replies teach it across model families, and slots pasted unchanged into other models, one of them ten times larger, act through the receiving model's own refusal direction and, where they transfer, give more harm-grounded refusal than a plain instruction to refuse.
Given examples of a backdoor firing, the word that independent searches share is a true trigger word for eight of twelve backdoors, and searching against refusal metrics shows what they miscount: a harm-grounded judge reads $0.806$ on declines of material that a slot injects, where hand labels give $0.442$.
\end{abstract}

\section{Introduction}
\label{sec:intro}

A behavior can be put into a language model by changing its internals, by adding a steering vector \citep{turner2023actadd,rimsky2024caa}, clamping a sparse-autoencoder (SAE) feature \citep{lieberum2024gemmascope} or fine-tuning in a backdoor \citep{hubinger2024sleeper}, or by changing only its prompt.
The two differ in the internal state they produce: almost surely no prompt reproduces a steered state \citep{mishra2026nonsurjective}, and in two function-vector settings steering changes activations by $26$ and $77$ times the median natural change \citep{shih2026steeringcomputes}.
They are also related.
Refusal is largely mediated by one direction in activation space, and jailbreaks and adversarial suffixes work in part by suppressing it \citep{arditi2024refusal,ball2024jailbreaksuccess}.
What has not been tested is whether prompt tokens found by imitating a modification depend on that modification's direction, and whether a plain instruction for the same behavior does.

\begin{figure*}[t]
\centering
\resizebox{\textwidth}{!}{
\begin{tikzpicture}[
    font=\small, >={Stealth[length=1.9mm]},
    m/.style={draw=black!45, rounded corners=1.5pt, fill=white, minimum height=0.7cm, align=center,
              inner sep=3pt},
    ma/.style={m, draw=ttaccent, fill=ttaccent!10, line width=0.7pt},
    tok/.style={draw=ttaccent, fill=ttaccent!14, minimum width=0.34cm, minimum height=0.34cm, inner sep=0pt},
    tokx/.style={draw=black!35, fill=black!6, minimum width=0.34cm, minimum height=0.34cm, inner sep=0pt},
    arr/.style={->, draw=black!60, line width=0.65pt},
    arra/.style={->, draw=ttaccent, line width=0.95pt},
    ttl/.style={font=\small\bfseries, anchor=west, inner sep=0pt},
    note/.style={font=\footnotesize, text=black!65, align=center, inner sep=1pt},
    sep/.style={draw=black!15, line width=0.5pt}]

  \node[ttl] at (0,3.3) {(a) Token steering};
  \node[ma, minimum width=1.9cm] (av)  at (1.0,2.4)  {model $+\,v$};
  \node[m,  minimum width=1.5cm] (ab)  at (3.4,2.4)  {behavior};
  \draw[arr] (av) -- (ab);
  \node[tok] (t1) at (0.64,1.25) {}; \node[tok] (t2) at (1.0,1.25) {}; \node[tok] (t3) at (1.36,1.25) {};
  \node[note, anchor=east] at (0.42,1.25) {$k$ tokens};
  \node[m,  minimum width=1.9cm] (am)  at (1.0,0.3)  {model};
  \node[m,  minimum width=1.5cm] (ab2) at (3.4,0.3)  {same\\[-2pt]behavior};
  \draw[arr] (t2) -- (am);
  \draw[arr] (am) -- (ab2);
  \draw[arra, dashed] (ab.south) .. controls (3.4,1.45) .. (t3.east);
  \node[note, fill=white] at (2.75,1.55) {search};

  \draw[sep] (4.55,-0.2) -- (4.55,3.5);

  \begin{scope}[xshift=4.85cm]
  \node[ttl] at (0,3.3) {(b) Route test};
  \node[note, anchor=west, text=black] at (0,2.55) {slot, nothing removed};
  \node[note, anchor=west, text=black] at (0,1.85) {slot, $v$ removed};
  \node[note, anchor=west, text=black] at (0,1.15) {slot, random dir.\ removed};
  \node[note, anchor=west, text=black, align=left] at (0,0.3) {new search,\\[-2pt]$v$ removed};
  \foreach \y in {2.55,1.85,1.15,0.3} { \draw[arr] (3.55,\y) -- (3.95,\y); }
  \node[note, anchor=west, text=ttkeep] at (4.0,2.55) {behavior};
  \node[note, anchor=west, text=ttcut]  at (4.0,1.85) {gone: needs $v$};
  \node[note, anchor=west, text=ttkeep] at (4.0,1.15) {kept (control)};
  \node[note, anchor=west, text=black!80] at (4.0,0.3) {found without $v$?};
  \end{scope}

  \draw[sep] (11.45,-0.2) -- (11.45,3.5);

  \begin{scope}[xshift=11.75cm]
  \node[ttl] at (0,3.3) {(c) Moving a behavior};
  \node[ma, minimum width=2.0cm] (ct) at (1.05,2.4) {model A $+\,v_A$};
  \node[m,  minimum width=1.6cm] (cr) at (4.05,2.4) {replies\\[-2pt]as text};
  \draw[arr] (ct) -- (cr);
  \node[m,  minimum width=2.0cm] (cs) at (1.05,0.3) {model B $+$ slot};
  \node[m,  minimum width=1.6cm] (cc) at (4.05,0.3) {cut B's own $v_B$};
  \draw[arra, dashed] (cr.south) .. controls (4.05,1.35) and (1.05,1.35) .. (cs.north);
  \node[note, fill=white] at (2.55,1.35) {search on B};
  \draw[arr] (cs) -- (cc);
  \end{scope}

  \draw[sep] (17.2,-0.2) -- (17.2,3.5);

  \begin{scope}[xshift=17.5cm]
  \node[ttl] at (0,3.3) {(d) Audit};
  \foreach \y/\h in {2.6/2, 2.15/1, 1.7/3, 1.25/2} {
    \foreach \x in {1,2,3} {
      \ifnum\x=\h \node[tok] at (0.05+0.4*\x,\y) {}; \else \node[tokx] at (0.05+0.4*\x,\y) {}; \fi
    }
  }
  \node[note, anchor=north] at (0.85,1.0) {independent searches};
  \node[m, minimum width=1.25cm] (ds) at (2.65,1.92) {shared\\[-2pt]token};
  \draw[arr] (1.45,1.92) -- (ds.west);
  \node[m, minimum width=2.1cm] (df) at (2.3,0.3) {insert: fires?};
  \draw[arr] (ds.south) -- (ds.south |- df.north);
  \end{scope}
\end{tikzpicture}%
}
\caption{Token steering and its uses. (a) A search finds $k$ prompt tokens that lead the unmodified model to behave as it does with the modification $v$ added. (b) The route test removes $v$, or a random direction as the control, and scores the slot again; a search on the model without $v$ looks for a slot that works without it. (c) The search can instead target another model's replies, saved as text; the direction removed is then the receiving model's own. (d) Independent searches on a backdoored model share the trigger word, which is then inserted into fresh prompts to see whether the backdoor fires.}
\label{fig:overview}
\end{figure*}
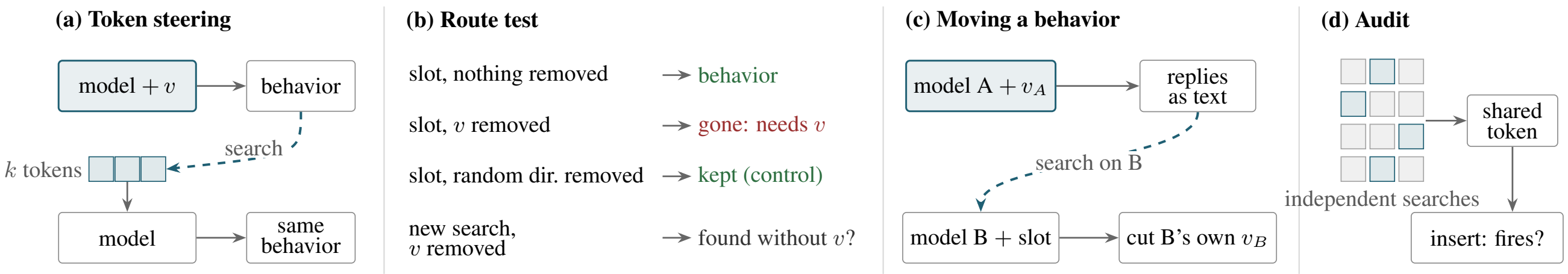

Prior work optimizes prompt tokens toward an internal target, such as a feature, a neuron or a direction \citep{mody2026minimizing,cakar2026optimizing}; there the tokens act on the target by construction.
We use the tokens as an instrument instead.
\emph{Token steering} searches for a few prompt tokens, a \emph{slot}, that make an unmodified model behave like its modified counterpart.
The search sees only the modified model's behavior, never the direction of the modification.
The \emph{route test} then removes that direction from the model and scores the slot again, so it can come out either way: if the behavior disappears, the behavior the slot produces needs the direction (Figure~\ref{fig:overview}).

Our main finding is that searched slots need the direction they were never shown, while plain instructions for the same behavior often keep working without it.
We establish it on refusal (Section~\ref{sec:core}).
On gemma-2b-it, Llama-3-8B-Instruct and Yi-6B-Chat, three searched tokens make the model refuse harmless requests, as a published refusal vector does \citep{arditi2024refusal}.
Removing the refusal direction takes all $62$ slots we judged to $0.004$ or less on a harm-grounded judge, and removing a random direction leaves them in place.
A search for short slots run without the direction reaches at most $0.19$, while the same search with a random direction removed scores $0.94$--$1.03$ of the ordinary search on gemma-2b-it.
Plain instructions behave differently.
With the direction removed, instructions written to state a harm reason keep most of their harm-grounded refusal on Llama-3-8B-Instruct (a mean of $0.775$ falls to $0.604$), and on gemma-2b-it plain instructions keep a plain decline where slots lose theirs ($0.46$ to $0.28$ against $0.80$ to $0.03$ on a substring check).

The method also shows, or makes possible, four further things:
\begin{itemize}
\item \textbf{The same dependence for other behaviors} (Section~\ref{sec:general}). Slots that imitate a Spanish SAE feature, a Spanish steering vector, three function vectors, a French backdoor and an evil persona vector need their direction; for all but the persona, plain instructions need it too.
\item \textbf{Refusal moved between models through text} (Section~\ref{sec:transfer}). Slots searched against a steered model's saved replies teach refusal across model families, and removing the receiving model's own direction removes $99$--$100\%$ of it. Slots pasted unchanged into models no search ran on give more harm-grounded refusal than a plain instruction to refuse: $0.883$ against $0.280$ on gemma-2-9b-it ($0.750$ against $0.217$ on hand labels, $n=60$ per condition), and $0.370$ against $0.017$ on a model ten times larger than the slot's donor.
\item \textbf{An audit for backdoor triggers} (Section~\ref{sec:audit}). Given examples of a backdoor firing, the word that independent searches share is a true trigger word for eight of twelve public backdoors.
\item \textbf{A view of refusal metrics} (Section~\ref{sec:measurement}). A search aimed at a refusal metric finds the replies the metric miscounts: on declines of material that a slot injects, the harm-grounded judge reads $0.806$ where hand labels give $0.442$.
\end{itemize}

\section{Token Steering and the Route Test}
\label{sec:method}

\paragraph{Token steering.}
We start from a \emph{good model} and a \emph{bad model}.
For a steering vector, an SAE feature or a function vector, they are the unmodified model and the same model with the modification; for a backdoored fine-tune, the fine-tune on a clean prompt and on a prompt with its trigger.
A \emph{slot} is a sequence of $k$ tokens placed before (prefix) or after (postfix) an instruction, and a \emph{cell} is one combination of model, position, slot length and seed.
Token steering searches for a slot that makes the good model behave like the bad one, with the tokens as actions and one of two rewards: an \emph{imitation} reward, the mean log-likelihood the good model, given the slot, assigns to the bad model's greedy replies; or a \emph{behavior} reward, a detector applied to the good model's own reply.
Neither reward sees the modification's direction.
We run eight search techniques and make no claim about which is better (Appendix~\ref{app:search}).

\paragraph{The route test.}
For a modification with direction $v$, the route test removes $v$ from the good model's activations at every layer and position, including those of the reply, and rescores the slot.
With $s$ the slot's score intact, $s_{\text{abl}}$ its score with $v$ removed and $s_0$ the model's score with no slot, the \emph{mediated fraction}
\begin{equation}
\mf = \frac{s - s_{\text{abl}}}{s - s_0}
\end{equation}
is the share of the slot's effect that disappears with the direction.
We say that a behavior \emph{needs} a direction when it disappears with the direction removed.
Because $v$ is removed from the whole model, this is a statement about the behavior: it does not show that the slot acts by writing to $v$, and it would also hold if the model could not produce the behavior at all without $v$.
The test therefore tells slots and instructions apart only where an instruction keeps the behavior under the same removal, which we find in three cases: harm-grounded refusal on Llama-3-8B-Instruct, a plain decline on gemma-2b-it, and the evil persona.
We did not test removal restricted to the slot's positions, nor sufficiency, that is, whether restoring only the slot's push along $v$ restores the behavior.

Refusal directions are zero-ablated (projected out), as in \citet{arditi2024refusal}.
Backdoor, contrastive and persona directions are mean-ablated, their component set to its average on clean prompts \citep{marshall2024affine}, because zeroing them changes the model with nothing in the prompt (Section~\ref{sec:general}).
A backdoor's direction is the difference of mean activations between triggered and clean prompts.
For refusal $s_0$ is at most $0.03$, so we set $s_0 = 0$, and we read a slot's mediated fraction only when $s - s_0$ is at least $0.10$.

\paragraph{Controls.}
Three controls guard the reading of $\mf$ (Figure~\ref{fig:overview}b).
The \emph{random-direction cut} removes a random direction in the same way and should leave the behavior in place.
The \emph{ablated search} reruns the search on the model with $v$ removed throughout, to look for a slot that produces the behavior without $v$.
The \emph{search control} runs the same search with a random direction removed, to check that a failed ablated search is not the search breaking.

\paragraph{Moving a behavior.}
To \emph{teach} a behavior to another model, we save a steered model's replies as text, encode them with the receiving model's tokenizer, and search on the receiver with the imitation reward.
To \emph{paste} it, we place a slot found on one model, unchanged, in the prompt of a model no search ran on, next to a random slot, harm-related words of the same length and plain instructions to refuse.
In both cases the route test removes the receiver's own direction.

\paragraph{Instruments.}
The \emph{harm-grounded judge} is Qwen2.5-32B-Instruct \citep{yang2024qwen25}, which sees only the request and the reply and answers a yes-or-no rubric that counts only declines grounded in harm, law, ethics or policy.
It is our default: every refusal rate without a tag is the judge's.
The \emph{substring check} is the published list of refusal phrases \citep{arditi2024refusal}; it counts a reply as a refusal if it contains any of them, with or without a reason.
The judge agrees with hand labels on $0.717$--$0.967$ of items across $12$ models (Section~\ref{sec:measurement}).

\paragraph{Setup.}
The core refusal experiments use gemma-2b-it, Llama-3-8B-Instruct and Yi-6B-Chat with the refusal directions published by \citet{arditi2024refusal}.
Four further chat models (gemma-2-9b-it, Llama-3.1-8B-Instruct, Qwen2.5-7B-Instruct and Llama-2-7b-chat) and a reasoning model (Qwen3-8B) extend them; where no direction is published we extract one with the same recipe (Appendix~\ref{app:models}).
The other modifications are a Spanish SAE feature \citep{lieberum2024gemmascope} and contrastive Spanish and persona vectors \citep{rimsky2024caa} on gemma-2-2b-it, function vectors on GPT-J-6B \citep{todd2024function}, and public backdoored fine-tunes.
Refusal is induced on harmless instructions: slots are searched on $32$ of them and scored on $100$--$500$ held-out ones.
Table~\ref{tab:grid} lists every modification the route test was run on and what it found.

\begin{table}[t]
\caption{The route test on every modification we tried: what removing the modification's direction does to the behavior that searched slots and plain instructions produce, and what a search run without the direction finds. ``Lost'': the effect disappears; ``part'': part of it is kept; a dash: not run. Sections~\ref{sec:core}--\ref{sec:transfer} and Appendices~\ref{app:core}--\ref{app:transferfull} give the numbers behind each cell.}
\label{tab:grid}
\centering
\small
\setlength{\tabcolsep}{3pt}
\begin{tabular}{@{}L{3.05cm}L{1.55cm}L{0.95cm}L{1.9cm}@{}}
\toprule
Modification & Slots & Instr. & Search \\
\midrule
Refusal vector, 3 core models & \cellLost{lost} & \cellPart{part} & \cellLost{at most $0.19$} \\
Refusal vector, 4 further models & \cellPart{lost on 3} & \cellNA{---} & \cellPart{$0.003$ or less on 3} \\
Refusal taught across models & \cellPart{lost on all but 1} & \cellNA{---} & \cellPart{part on the 1 run} \\
Evil persona vector & \cellLost{lost} & \cellPart{part} & \cellNA{---} \\
Spanish SAE feature & \cellLost{lost} & \cellLost{lost} & \cellLost{none found} \\
Spanish contrastive vector & \cellLost{lost} & \cellLost{lost} & \cellNA{---} \\
Function vectors, 3 tasks & \cellLost{lost} & \cellLost{lost} & \cellLost{none found} \\
French backdoor, 2 models & \cellLost{lost} & \cellLost{lost} & \cellLost{none found} \\
Flattery vector & \cellNA{not reached} & \cellPart{part} & \cellNA{---} \\
Refusal and hate-message backdoors (6) & \multicolumn{3}{>{\raggedright\arraybackslash}p{\dimexpr 4.4cm+4\tabcolsep\relax}}{\cellNA{no single direction to remove}} \\
\bottomrule
\end{tabular}
\end{table}

\section{Searched Refusal Slots Need the Refusal Direction}
\label{sec:core}

On the three core models, three searched tokens make the model refuse harmless requests, and that refusal needs the refusal direction: it disappears when the direction is removed and stays when a random direction is removed.
Plain instructions to refuse depend on the direction only partly.

\begin{figure*}[t]
\centering
\includegraphics[width=\textwidth]{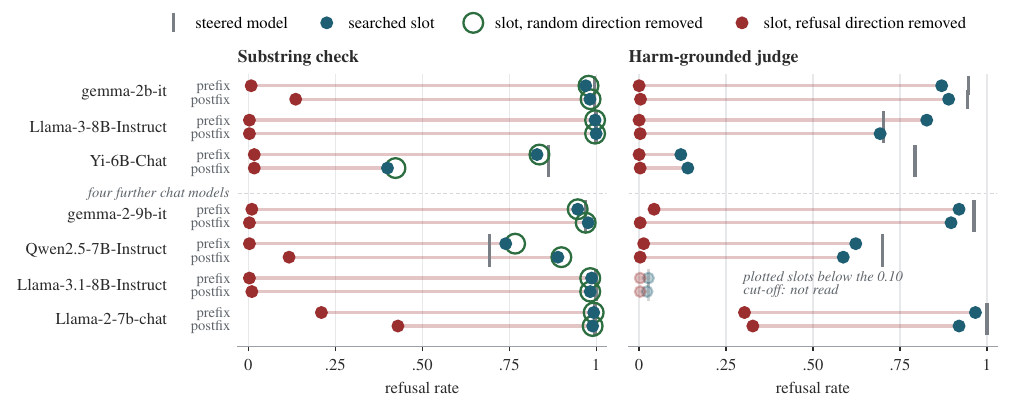}
\caption{Removing the refusal direction removes token-steered refusal; removing a random direction does not. Rows are models and prompt positions; marks are the steered model (tick), the best three-token slot (blue), and that slot with the refusal direction removed (red) or a random direction removed (green ring). The substring check (left) scores all four conditions; the harm-grounded judge (right) was not run on the random-direction cut. Llama-3.1-8B-Instruct's plotted slots are faded because they read below the $0.10$ at which we read a mediated fraction; its two slots that read highest on the judge, at $0.100$ and $0.220$, fall to $0.000$ (numbers in Table~\ref{tab:core}, Appendix~\ref{app:core}).}
\label{fig:core}
\end{figure*}

\paragraph{Three tokens reproduce the vector.}
On gemma-2b-it the best three-token slots reach $0.870$ at the start of the prompt and $0.890$ at the end, against $0.948$ and $0.944$ for the steered model (Figure~\ref{fig:core}).
On Llama-3-8B-Instruct the $24$ judged slots reach $0.35$--$0.88$, where the vector scores $0.70$.
On Yi-6B-Chat the slots reproduce only part of the vector: the plotted slots reach $0.12$ and $0.14$, against $0.79$.
Random slots of the same length do nothing: over $108$ judged readings their median is $0.001$, and the judge reads no random slot in our refusal runs above $0.027$.

\paragraph{Removing the direction removes the refusal.}
With the refusal direction removed, all $62$ judged slots, $14$ on gemma-2b-it and $24$ each on Llama-3-8B-Instruct and Yi-6B-Chat, fall to $0.004$ or less.
A random direction does not do this.
In Figure~\ref{fig:core} the random-direction control was scored on the substring check, not the judge: there a random direction leaves each plotted slot of the core models within $0.03$ of its own rate, while the refusal direction takes them to $0.003$--$0.136$.

\paragraph{The result holds on other prompts, models and slots.}
Scored with no new search on a second prompt set, the context-free prompts of \texttt{databricks-dolly-15k}, the best slots reach $0.893$ and $0.887$ on gemma-2b-it (vector $0.957$) and $0.903$ and $0.653$ on Llama-3-8B-Instruct (vector $0.840$); all four fall to $0.000$ with the direction removed, while three random cuts leave each within about $0.03$ of its rate.
The result holds on three of four further chat models; on the fourth, Llama-2-7b-chat, the slots keep about a third of their refusal ($31\%$ and $36\%$; Appendix~\ref{app:more}).
It holds on a reasoning model: on Qwen3-8B, run with its thinking block closed, the best three-token slot falls from $0.16$ to $0.00$, while three random directions leave it at $0.19$--$0.23$.
It also holds for $32$-token slots searched, at three seeds on Llama-3.1-8B-Instruct and Qwen2.5-7B-Instruct, with $38$ decline and harm words banned.
Every such slot reads $0.000$ with the direction removed, while three random directions leave each slot that reads above $0.05$ with at least $72\%$ of its rate.
In $8$ of the $12$ seed and position cells, the best slot free of every banned word, at $0.155$--$0.437$, also beats the better of two plain instructions to refuse (Appendix~\ref{app:core} gives the ban's limits).

\begin{figure}[t]
\centering
\includegraphics[width=\columnwidth]{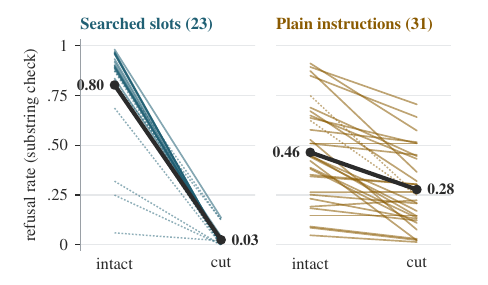}
\caption{Slots and instructions on gemma-2b-it, intact and with the refusal direction removed (``cut''), on the substring check. Each line is one slot or one instruction and thick lines are means. Dotted lines were scored on $16$ or $50$ prompts, solid ones on $500$. One instruction that rises with the cut ($0.492$ to $0.600$) is left out; with it the instructions' means are $0.47$ and $0.29$.}
\label{fig:routecut}
\end{figure}

\paragraph{Plain instructions depend on the direction only partly.}
Hand-written instructions to decline keep part of their effect without the direction, and what they keep differs by model.
On gemma-2b-it they keep a plain decline: removing the direction takes searched slots from $0.80$ to $0.03$ on average, about $97\%$ of their effect, and plain instructions from $0.46$ to $0.28$, about $40\%$ (substring check; Figure~\ref{fig:routecut}); $22$ of $23$ slots lose at least $80\%$ of their effect, against $2$ of $31$ instructions.
The decline they keep has no harm reason: on the judge, the best instruction falls from $0.49$ to $0.06$.
On Llama-3-8B-Instruct instructions keep the harm reason.
Few ordinary instructions make this model give one, so we wrote $15$ that state a harm reason; removing the direction takes their mean rate from $0.775$ to $0.604$ at the start of the prompt and from $0.619$ to $0.512$ at the end, a median loss of $3\%$ and $4\%$, while the slots lose all of theirs.
On Yi-6B-Chat instructions mostly need the direction (median $\mf$ $0.56$ and $0.76$ on a broad rubric; Appendix~\ref{app:core}).

\begin{figure}[t]
\centering
\includegraphics[width=\columnwidth]{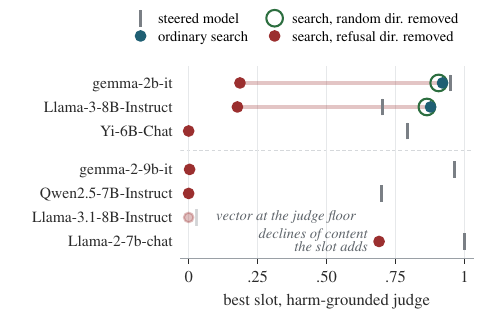}
\caption{Searching without the refusal direction (best slot over both positions, seed 0). Red: the best slot from a search run with the refusal direction removed, scored with it removed; on gemma-2b-it, the highest of $22$ cells at three and ten tokens. Blue and green: the ordinary search, and the same search run with a random direction removed. Ticks mark the steered model.}
\label{fig:search}
\end{figure}

\paragraph{Searching without the direction.}
A search for short slots finds little harm-grounded refusal without the direction.
We reran the search, for slots of three tokens and on gemma-2b-it also of ten, on each model with its refusal direction removed throughout (Figure~\ref{fig:search}).
The best slot reaches $0.19$ on gemma-2b-it, $0.18$ on Llama-3-8B-Instruct and $0.00$ on Yi-6B-Chat.
The search itself still works: run with a random direction removed instead, it reaches $0.94$--$1.03$ of the ordinary search's score on gemma-2b-it at three seeds and $0.96$ and $0.98$ on Llama-3-8B-Instruct, and the slots it finds fall below $0.01$ when the refusal direction is removed.
On three further chat models the search reaches $0.003$ or less.
It finds more on two models.
On Llama-3.2-3B it finds part of a route: its best slots reach $0.257$ and $0.313$, against $0.460$ for the vector.
On Llama-2-7b-chat it finds declines that survive, at $0.620$--$0.690$, which a model-assisted check reads as aimed at content the slot itself adds (Appendix~\ref{app:more}).
These searches looked for short slots, so they do not show that no prompt works without the direction: on Llama-3-8B-Instruct the instructions above, $6$--$14$ words long, keep their harm-grounded refusal.

\section{Beyond Refusal}
\label{sec:general}

Wherever a search finds tokens that reproduce a modification, the tokens need its direction.
What changes from one behavior to the next is whether plain instructions need it too (Table~\ref{tab:beyond}).

\begin{table}[t]
\caption{The route test beyond refusal. ``Slot'' is the best searched slot and ``Instr.'' a plain instruction; ``removed'' removes the modification's direction (z: zero ablation, m: mean ablation) and ``random'' a random direction. Scores are the judge's rubric for Spanish and the persona, the backdoor's own detector, and task accuracy for the function vectors, on $200$, $100$, $128$ and $41$--$317$ test prompts. The two Spanish rows are the same slot and instruction with two different directions removed. Appendix~\ref{app:general} gives seeds and how each slot was chosen.}
\label{tab:beyond}
\centering
\small
\setlength{\tabcolsep}{2pt}
\begin{tabular}{@{}lccccc@{}}
\toprule
 & \multicolumn{3}{c}{Slot} & \multicolumn{2}{c}{Instr.} \\
\cmidrule(lr){2-4}\cmidrule(l){5-6}
Modification & intact & removed & random & intact & removed \\
\midrule
\multicolumn{6}{@{}l}{\emph{gemma-2-2b-it}} \\
Spanish SAE feature$^{\mathrm{z}}$ & $0.930$ & $0.000$ & $0.935$ & $0.945$ & $0.000$ \\
Spanish vector$^{\mathrm{m}}$ & $0.930$ & $0.000$ & $0.935$ & $0.945$ & $0.000$ \\
Evil persona$^{\mathrm{m}}$ & $0.520$ & $0.000$ & $0.500$ & $0.940$ & $0.270$ \\
\multicolumn{6}{@{}l}{\emph{French backdoor$^{\mathrm{m}}$}} \\
gemma-2-2b & $1.000$ & $0.016$ & $1.000$ & $0.953$ & $0.008$ \\
Llama-3.2-3B & $0.961$ & $0.023$ & $0.977$ & $0.977$ & $0.055$ \\
\multicolumn{6}{@{}l}{\emph{Function vectors$^{\mathrm{z}}$, GPT-J-6B}} \\
antonym & $0.609$ & $0.000$ & $0.621$ & $0.612$ & $0.000$ \\
capitalize & $0.341$ & $0.006$ & $0.341$ & $0.618$ & $0.000$ \\
country-capital & $0.854$ & $0.000$ & $0.854$ & $1.000$ & $0.024$ \\
\bottomrule
\end{tabular}
\end{table}

\paragraph{Slots and instructions both need the direction.}
For a Spanish SAE feature, a contrastive Spanish vector, three function vectors and a French backdoor, removing the direction stops searched slots and plain instructions alike, so here the test shows that the behavior needs the direction and does not separate the two.
The best Spanish slot falls from $0.930$ to $0.000$ and a plain instruction to answer in Spanish from $0.945$ to $0.000$.
On GPT-J-6B the best slots for three function vectors reach $0.341$--$0.854$ task accuracy and fall to $0.006$ or less, and instructions that name the task fall from $0.612$--$1.000$ to $0.024$ or less.
On a backdoored gemma-2-2b that answers in French when triggered, the best slot falls from $1.000$ to $0.016$ on the backdoor's detector and a plain instruction from $0.953$ to $0.008$.
A random direction leaves the best slot of each cell within $0.02$ of its rate, and the same reading holds at two more search seeds.
A search run without the direction finds nothing: $0.01$ or less on Spanish, $0.00$ on the antonym task and $0.016$ or less on the French backdoor.
In six of the seven distinct comparisons in Table~\ref{tab:beyond}, the instruction reaches the behavior at least as well as the best slot.

\paragraph{The evil persona separates them.}
The evil persona vector behaves like refusal.
It makes gemma-2-2b-it give evil replies at $0.52$ on the judge's evil rubric, and slots of $32$--$64$ tokens reproduce it.
With the evil direction removed, all $10$ slots tested fall from $0.30$--$0.52$ to $0.00$--$0.02$, while a random direction leaves $0.80$--$1.04$ of their effect; plain instructions keep $25$--$36\%$ of theirs.

\paragraph{Flattery: instructions reach it, our search did not.}
No slot we found scores above $0.405$ on the judge's sycophancy rubric, where the flattery vector scores $0.625$.
Plain commands do reach flattery, and they differ in how much they need the direction: the command the vector was built from loses $91\%$ and $94\%$ of its effect with the direction removed, and a reworded command $9\%$ and $57\%$.

\paragraph{Removing a direction can install a behavior.}
With nothing in the prompt, zero-ablating the contrastive Spanish direction makes gemma-2-2b-it answer in Spanish at $0.955$, while mean ablation, a random direction and the plain model all stay at $0.000$.
This is why we mean-ablate the backdoor, contrastive and persona directions; anyone who removes a direction to switch a behavior off should first check the model with nothing in the prompt.

\paragraph{Where the test could not be run.}
Six backdoors whose payload is a refusal or a hate message have no single direction to remove: the best candidate takes away $0$--$27\%$ of the trigger's effect.
Four further persona vectors (corrigible, survival, myopic and deception) score $0.00$--$0.01$ on their own rubrics, so they give no behavior to copy.
A vector meant to make Qwen3-1.7B think longer was not copied: searched slots add $0.03$--$0.06$ of the vector's gain (Appendix~\ref{app:general}).

\section{Moving Refusal Between Models}
\label{sec:transfer}

Refusal can be moved from one model to another through text alone, and on the receiving model it needs the receiver's own refusal direction.

\paragraph{Teaching across model families.}
Llama-3-8B-Instruct learns refusal from a steered gemma-2b-it, a model about a third its size, and gemma-2b-it learns it from a steered Llama-3-8B-Instruct.
At both prompt positions and two seeds, three-token slots reproduce $0.94$--$1.01$ of the teacher's refusal rate on the substring check, and removing the receiver's own direction removes $99$--$100\%$ of it (Table~\ref{tab:transfer}, Appendix~\ref{app:transferfull}).
On the other receivers where a mediated fraction can be read, removing the receiver's direction takes away $92$--$100\%$ of the taught refusal at the first seed.
The exception is Llama-3.2-3B, which keeps part of the refusal it is taught when its direction is removed ($\mf$ $0.54$ and $0.78$).
Over all receivers, three-token slots reproduce $0.49$--$1.01$ of the teacher, and longer slots raise five of the six low cells we reran to $0.89$--$1.00$.
Spanish, a French behavior taught from a backdoored fine-tune and, at one cell of four, an evil persona cross in the same way and need the receiver's own direction, but for them a plain instruction does as well as or better than any slot; emergent misalignment \citep{betley2025emergent} did not cross from any of three teachers (Appendix~\ref{app:transferfull}).

\paragraph{Pasting, with no search on the receiver.}
We paste $24$ refusal slots, $16$ found on gemma-2b-it and $8$ on Llama-3-8B-Instruct, into five models that no search ran on (Table~\ref{tab:paste}).
They transfer to four of the five, and on the two of those that have a refusal direction of their own, removing it removes most or all of the effect.
On gemma-2-9b-it the best pasted slot reaches $0.883$, against $0.973$ for the model's own vector; for the $22$ slots that reach $0.10$, removing the receiver's direction takes away $91$--$100\%$ of the effect, and random directions leave each slot at $0.91$--$1.14$ of its rate.
On Llama-3.2-3B the best reaches $0.687$, above the model's own vector at $0.533$, and for the $16$ slots that reach $0.10$ removing the direction takes away a median of $85\%$ of the effect ($55$--$100\%$).
On Yi-6B-Chat the transfer fails: the best slot reaches $0.15$.

\begin{table}[t]
\caption{Refusal slots pasted unchanged into models that no search ran on: the best pasted slot, the better of two plain instructions to refuse, harm-related words of the slot's length, the receiver's own refusal vector, and the share of the effect lost with the receiver's direction removed (the range over slots that reach $0.10$ on gemma-2-9b-it, their median on Llama-3.2-3B, the best slot's in the lower block). Upper block: slots of $3$ and $10$ tokens from gemma-2b-it and Llama-3-8B-Instruct; lower block: $32$-token slots from a smaller model of the same family. The best slot is picked on the prompts it is scored on; the median slot reads $0.393$, $0.175$, $0.905$ and $0.305$ on the first four receivers. Random slots read at most $0.027$.}
\label{tab:paste}
\centering
\small
\setlength{\tabcolsep}{2.4pt}
\begin{tabular}{@{}lccccc@{}}
\toprule
Receiver & Slot & Instr. & Harm & Vector & Lost\,\% \\
\midrule
gemma-2-9b-it   & $0.883$ & $0.280$ & $0.220$ & $0.973$ & $91$--$100$ \\
Llama-3.2-3B    & $0.687$ & $0.023$ & $0.190$ & $0.533$ & $85$ \\
gemma-1.1-2b-it & $1.000$ & $0.400$ & ---     & ---     & --- \\
gemma-7b-it     & $0.800$ & $0.040$ & ---     & ---     & --- \\
Yi-6B-Chat      & $0.15$  & ---     & $0.053$ & $0.58$  & --- \\
\midrule
Qwen2.5-72B     & $0.370$ & $0.017$ & $0.077$ & $0.857$ & $85$ \\
Llama-3.1-70B   & $0.190$ & $0.005$ & $0.205$ & $0.315$ & $100$ \\
Qwen2.5-14B     & $0.070$ & $0.070$ & $0.100$ & $0.347$ & --- \\
Qwen2.5-32B     & $0.040$ & $0.017$ & $0.033$ & $0.600$ & --- \\
\bottomrule
\end{tabular}
\end{table}

\paragraph{Pasted slots against a plain instruction.}
Where a pasted slot transfers, it produces more harm-grounded refusal than a plain instruction to refuse: $0.883$ against $0.280$ on gemma-2-9b-it and $0.687$ against $0.023$ on Llama-3.2-3B, where harm-related words of the slots' lengths reach at most $0.220$ and $0.190$.
The instructions do produce declines, but without a harm reason: on gemma-7b-it the same instruction reads $0.315$--$0.830$ on the substring check and $0.010$--$0.040$ on the judge.
The comparison is with plain instructions to refuse; instructions written to state a harm reason were tried only on Llama-3-8B-Instruct, where they do produce harm-grounded refusal (Section~\ref{sec:core}).
Because the judge can over-count declines of material that a slot adds (Section~\ref{sec:measurement}), one of the authors labeled $60$ replies drawn at random from each pasted condition on two receivers, blind to the judge.
On the labels the slot still beats the instruction: $0.750$ against $0.217$ on gemma-2-9b-it, where the judge reads the slot at $0.883$, and $0.417$ against $0.017$ on Qwen2.5-72B-Instruct, where the judge reads it at $0.367$.

\paragraph{A model ten times larger.}
A refusal slot searched on Qwen2.5-7B-Instruct works pasted into Qwen2.5-72B-Instruct.
It reaches $0.370$, against $0.017$ for a plain instruction, $0.077$ for $32$ harm-related words and $0.000$ for a random slot, and removing the larger model's refusal direction takes it to $0.057$ ($\mf$ $0.847$).
This result rests on one $32$-token slot at the end of the prompt.
On Llama-3.1-70B-Instruct the best pasted slot reaches $0.190$ and loses all of it with the direction removed, but harm-related words do as well as the slot ($0.205$); on Qwen2.5-14B and Qwen2.5-32B nothing transfers.
Answering in another language is the opposite case: in $115$ comparisons, no pasted language slot beats a plain instruction to answer in that language (Appendix~\ref{app:language}).

\section{Auditing for a Hidden Trigger}
\label{sec:audit}

Given examples of a backdoor firing, the word that independent searches share names its trigger.
The trigger is the cheapest way to set off a backdoor, so independent searches converge on it while the rest of each slot varies.

\paragraph{The statistic.}
We audit public backdoored fine-tunes whose triggers are documented.
Independent searches, with different searchers, seeds and positions, look for slots that make the backdoored model, on clean prompts, reproduce its own replies to triggered prompts or produce its payload; they never see the trigger.
We keep up to six slots per model that produce the payload on at least half of held-out clean prompts while leaving the clean base model at $0.2$ or below, and call the most-shared word the trigger when at least $0.40$ of the kept slots contain it.

\begin{table}[t]
\caption{The shared-word audit on $12$ public backdoors, given examples of the backdoor firing. ``Share'' is the fraction of kept slots that contain the most-shared word; a word is called at $0.40$ or more, and every called word is a true trigger word. ``Fires'' counts the replies, of $30$ fresh prompts with the called word inserted, that carry the payload (for a two-word backdoor, its two most-shared words are inserted). Appendix~\ref{app:audit} gives the controls.}
\label{tab:auditmain}
\centering
\small
\setlength{\tabcolsep}{3.5pt}
\begin{tabular}{@{}lllcc@{}}
\toprule
Model & Payload & Trigger & Share & Fires \\
\midrule
gemma-2-2b & refusal & two words & $1.00$ & $3$ \\
Qwen2.5-3B & hate message & one word & $0.67$ & $28$ \\
gemma-2-2b & hate message & two words & $0.67$ & $16$ \\
gemma-2-2b & French & one word & $0.67$ & $28$ \\
gemma-2-2b & French & two words & $0.50$ & $1$ \\
Qwen2.5-3B & refusal & one word & $0.50$ & $29$ \\
Llama-3.1-8B & hate message & one word & $0.50$ & $28$ \\
Llama-3.2-3B & French & one word & $0.50$ & $20$ \\
\midrule
gemma-2-2b & French & two words & $0.33$ & --- \\
Llama-3.2-3B & refusal & one word & $0.33$ & --- \\
Llama-3.2-3B & hate message & code-like & $0.17$ & --- \\
Qwen2.5-0.5B & hate message & code-like & $0.17$ & --- \\
\bottomrule
\end{tabular}
\end{table}

\paragraph{With examples.}
Of $12$ backdoors, the statistic calls $8$, each time with a true trigger word (Table~\ref{tab:auditmain}).
We chose the cut-off after seeing the slots it reads, so this count is not a held-out test of it.
Inserted before $30$ fresh prompts, the called word fires all five single-word backdoors among the eight: $20$ to $29$ of $30$ replies carry the payload with the word, none without it, and at most $2$ of $30$ on the clean base.
Of the three two-word backdoors called, one fires.
A blind LLM judge given only the kept slots and a checklist of trigger kinds names the trigger for $5$ of $9$ held-out models (Appendix~\ref{app:audit}).

\paragraph{Without examples.}
A payload-free variant rewards a slot for making the backdoored model less likely than its clean base to give the base's own reply, so it needs only the two models.
On five single-word backdoors it names two, where the statistic with examples calls four.
It names no wrong trigger on any of the nine models it was run on, two of them fine-tunes with no trigger, but its silence does not tell a clean model from a missed backdoor.

\paragraph{A one-token baseline.}
A vocabulary lookup that scores every token as a one-token slot \citep{hughes2026toxscreendetectingllmpoisoned} matches our search on single-word triggers: on both single-word backdoors we tried, it ranks a true trigger word first.
Where two words must appear together it fails, ranking the trigger words between $4{,}550$ and $42{,}329$ on the one such backdoor we tried; the search calls three of the four two-word backdoors.
The audit is therefore a forensic tool for an auditor who has already seen a backdoor fire, not a scanner for backdoors nobody has seen.

\section{What Searching Shows About Refusal Metrics}
\label{sec:measurement}

A search aimed at a refusal metric finds the replies that the metric miscounts.
Every instrument we tried has such a blind spot, the harm-grounded judge included.

\paragraph{Decoy declines.}
On gemma-2b-it with the refusal direction removed, a search rewarded by the substring check finds slots whose replies decline for missing information, most often by saying that the context does not contain the answer.
A \emph{broad rubric}, which counts any decline as a refusal, scores these replies at $0.65$ at three tokens and $0.80$ at ten with a 7B judge, and at $0.54$--$0.62$ with the 32B judge.
The harm-grounded judge, which we wrote after reading them, gives the three-token slots at most $0.004$.
On Llama-3-8B-Instruct the same search reaches $0.28$ on the broad rubric and $0.00$ on the harm-grounded one.
From these runs we built a benchmark of $837$ replies: $340$ decoys, $199$ harm-grounded refusals and $298$ replies that do not refuse.

\paragraph{Content-triggered declines.}
A check for a harm reason has its own blind spot.
We rewarded a $16$-token search on gemma-2b-it, with the direction removed, with a \emph{harm-word check}, which counts a decline only if it also contains a harm or safety word.
The search learned to bring objectionable material into the prompt, so that the model declines the material and not the request.
The harm-word check reads the two best slots at $0.97$ and $0.72$; the harm-grounded judge reads them at $0.12$ and $0.04$.

\paragraph{The judge over-counts them too.}
The harm-grounded judge over-counts declines aimed at material that a slot adds.
From five runs on three models we collected $2{,}990$ such replies; of the $2{,}987$ that have a judge verdict, the harm-word check counts $97.6\%$ as harm-grounded refusals and the judge $80.6\%$.
A person labeled $60$ of them: the judge agrees on $0.650$ ($\kappa$ $0.300$), over-counting $15$ refusals and missing $6$.
Reweighted to the whole set, the rate by hand is $0.442$ ($95\%$ CI $0.295$--$0.589$), so the judge overstates it by $0.364$.
We therefore quote the judge on such replies only next to hand labels, as for the pasted slots of Section~\ref{sec:transfer}, where the judge reads about $0.13$ above the labels on gemma-2-9b-it and $0.05$ below them on Qwen2.5-72B.

\paragraph{The judge against hand labels.}
On replies without injected material, the judge agrees with hand labels on $0.717$--$0.967$ of items across twelve models (Table~\ref{tab:gold}, Appendix~\ref{app:judge}); most label sets began as a blind pass by a language model and were then reviewed by a person.
On seven of the twelve it misses no harm-grounded refusal and on the other five at most $6$ in $120$ items, so a low rate is not an undercount.
Its errors are mostly over-counts, largest on the two later OLMo-2 post-training stages, where it reports about twice the rate by hand.
The substring check errs in both directions: on Qwen3-8B it reads the searched slots $2.6$ to $4.4$ times higher than the judge, and on the slot pasted into Qwen2.5-72B it reads $0.063$ where the judge reads $0.370$ and hand labels $0.417$.

\section{Related Work}
\label{sec:related}

\paragraph{Steering and prompting.}
Activation steering adds a direction to the residual stream \citep{turner2023actadd,zou2023repe,rimsky2024caa}, and task vectors compress a prompt into a direction \citep{hendel2023taskvectors,todd2024function}.
The two are not equivalent: almost surely no discrete prompt reaches a steered state \citep{mishra2026nonsurjective}, and function vectors are far larger than natural activation changes \citep{shih2026steeringcomputes}.
We study behavior, not state, and the route test is a mediation test in the sense of \citet{vig2020causal}.

\paragraph{Refusal directions.}
Refusal is largely mediated by one direction \citep{arditi2024refusal}, which jailbreaks and adversarial suffixes suppress \citep{ball2024jailbreaksuccess,cakar2026optimizing}; later work finds several independent refusal directions \citep{wollschlager2025geometry}.
Our search without the direction is an input-side test of that finding.

\paragraph{Searching for tokens.}
Our searchers come from discrete prompt optimization \citep{shin2020autoprompt,wen2023hardprompts,zou2023universal,liu2024autodan,sadasivan2024beast}.
Work that optimizes prompts toward an internal target makes finding the tokens the result \citep{mody2026minimizing,cakar2026optimizing}; here the target is a modified model and the tokens are an instrument for an ablation test.

\paragraph{Backdoors, transfer and scoring.}
Our backdoors are public fine-tunes in the style of sleeper agents \citep{hubinger2024sleeper}, and trigger inversion recovers a trigger from a model \citep{li2025invertingtrojans,shen2025bait}.
Adversarial suffixes transfer between models \citep{zou2023universal}; we ask whether the transferred behavior needs the receiver's own direction.
Refusal benchmarks measure refusals of benign requests \citep{rottger2024xstest,cui2025orbench}, and jailbreak scorers can be fooled by replies that comply in name only \citep{souly2024strongreject}; decoy declines are the version of this problem that optimization produces.
Appendix~\ref{app:related} discusses further related work.

\section{Conclusion}
\label{sec:conclusion}

Token steering finds a few prompt tokens that make an unmodified model behave like a modified one, and the route test removes the modification's direction to see whether the behavior needs it.
For searched tokens it does: for refusal on the three core models, three of four further chat models and a reasoning model, and for Spanish, function vectors, a French backdoor and an evil persona.
Plain instructions often keep the behavior without the direction.
The same tokens move refusal between models, where it needs the receiving model's own direction, name backdoor triggers for an auditor who has seen the backdoor fire, and expose what refusal metrics miscount.
A plain instruction is usually the better tool for producing a behavior; the searched slot is the better probe of what a direction does, because an instruction can produce the behavior in more than one way.

\paragraph{Limitations.}
The route test removes a direction from the whole model, so it shows that a behavior needs the direction, not how the tokens use it and not that the direction is enough; removal restricted to positions or layers is untested.
The four further refusal models and the reasoning model are single-seed, taught receivers are at most $9$B, and slots pasted into much larger models work through few slots and on two of four receivers.
The search needs model weights, the audit needs the clean base and examples of the backdoor firing, and every modification we study was made on purpose (Appendix~\ref{app:limitations}).

\section*{AI Use Statement}
AI coding assistants (Anthropic's Claude) wrote much of the experiment and analysis code under the authors' direction and helped draft and revise the text; the authors checked the reported numbers against the run outputs.
Language models are also part of the method, as stated where each is used: the harm-grounded judge, the blind first pass of most hand-label sets, reviewed by a person (Table~\ref{tab:gold}), the model-assisted check of the Llama-2-7b-chat replies (Section~\ref{sec:core}), and the blind judge of the audit (Section~\ref{sec:audit}).
The $420$ hand labels of the pasted conditions (Section~\ref{sec:transfer}) were made by one of the authors, with no model pass.

\section*{Impact Statement}
Every request we search or score on is harmless; harmful requests appear only in the published sets used to compute refusal directions.
Most of our searches look for tokens that make a model refuse such requests or answer in another language.
Some elicit harmful content: an evil persona, emergent misalignment, and the payloads of public backdoors, including a hate message.
These slots give no capability that a plain prompt or a published trigger does not: for the persona, a plain instruction does better than any slot we found, emergent misalignment did not cross at all, and the backdoors and their triggers are already public.
We searched only open-weight models, and the one search for tokens that make a model comply despite a refusal vector, on harmless requests, reached $0.00$.
We quote no slot and no model reply.
The decoy benchmark, the hand labels, the scoring code and the run outputs are provided as supplementary material; Appendix~\ref{app:ethics} describes them and the content warning that one part carries.
The audit is meant for an auditor who has already seen a backdoor fire, and we report its weak result without examples, two of five, so that nobody relies on it as a scanner.

\bibliography{refs}
\bibliographystyle{icml2027}

\newpage
\appendix
\onecolumn

\section{Models, Directions and Data}
\label{app:models}

\paragraph{Models.}
Besides the models of Section~\ref{sec:method}, taught refusal uses Llama-3.2-3B-Instruct and three OLMo-2-1124-7B post-training stages (SFT, DPO, Instruct) \citep{olmo2024olmo2,lambert2024tulu3} as receivers, and pasted slots also go into gemma-1.1-2b-it, gemma-7b-it, Llama-3.1-70B-Instruct and Qwen2.5-14B, 32B and 72B-Instruct.
The language control uses Qwen2.5 donors from 1.5B to 32B parameters, and the three emergent-misalignment teachers are Qwen2.5-7B-Instruct fine-tunes \citep{turner2025modelorganisms}.
Qwen3-8B, Qwen3-14B and Qwen3-32B run with the thinking block closed, so that the reply comes directly; Qwen3-1.7B carries the thinking-length vector.
The backdoored models are public fine-tunes of gemma-2, Llama-3.x and Qwen2.5 models (Table~\ref{tab:auditmain}).

\paragraph{Directions.}
\citet{arditi2024refusal} published a refusal direction for gemma-2b, Llama-3-8B, Yi-6B and Llama-2-7B, and those models' own searches use it.
For the other models, and for Yi-6B when it receives pasted slots, we derive a direction with the same recipe; on gemma-2b-it and Llama-3-8B the recipe recovers the published layer and position and matches the published vectors at cosine similarity $0.995$ and $0.987$.
Another group has published directions for Llama-3.2-3B and Yi-6B at our layers and positions; we use them as a check, and on Llama-3.2-3B for the search with the direction removed (Appendix~\ref{app:core}).
Llama-3.2-3B's steered model reads $0.513$ on the judge with our direction in the taught runs, $0.533$ in the paste runs and $0.480$ with the borrowed direction; over every run at the default coefficient it reads $0.460$--$0.533$.
The Spanish feature is from Gemma Scope \citep{lieberum2024gemmascope}, and the Spanish and persona vectors are contrastive-activation-addition vectors \citep{rimsky2024caa}.
A backdoor comes with no direction, so we build one on the backdoored model: the average activation at the last prompt token over prompts with the trigger, minus the average over the same prompts without it.
We compute it at several layers, in most runs every second layer from $4$ to $18$, and keep the one whose removal, by mean ablation, most reduces how often the true trigger fires on a second set of prompts; the two sets have $64$ and $16$ prompts and are never used for testing.

\paragraph{Data.}
Slots are scored on held-out instructions from the harmless split of \citet{arditi2024refusal}, based on \citet{taori2023alpaca}: $500$ for the core gemma-2b-it runs, apart from two early runs in Figure~\ref{fig:routecut} on $16$ and $50$, and $200$--$300$ for transfer runs.
None of them is among the $32$ search prompts, before or after normalizing case and punctuation.

\section{Searchers, Rewards and Ablation}
\label{app:search}

\paragraph{Searchers.}
The eight searchers are gradient-guided coordinate search \citep{zou2023universal}, greedy and beam search \citep{sadasivan2024beast}, a genetic algorithm \citep{liu2024autodan}, hard-prompt projection \citep{wen2023hardprompts}, and three reinforcement-learning policies: PPO \citep{schulman2017ppo}, REINFORCE \citep{diao2022bdpl}, and GRPO with a small language model as the policy \citep{shao2024deepseekmath,deng2022rlprompt}.
Beam search, the genetic algorithm and the REINFORCE policy draw their candidates from a gradient shortlist, as coordinate search does.
Each run gives its searchers one budget of scored sequences.
Early exploratory runs tried other searchers, which we dropped; no figure or table draws on them.
We make no claim that any searcher beats another: at a matched reward, budget and seeds on the Spanish feature, every genetic-minus-gradient interval includes zero, and the one reinforcement-learning-minus-gradient interval that excludes zero has the reinforcement-learning side worse ($-0.15$, [$-0.27$, $-0.04$]).

\paragraph{Rewards.}
Most runs use $32$ search prompts and score each step on a random subset of $2$, $4$, $8$ or $16$ of them; the final score is on all $32$.
The imitation reward is the good model's log-likelihood, given the slot, of the bad model's greedy reply to each prompt (its first $32$ tokens for refusal), averaged over the replies; it has a gradient.
The behavior reward scores the good model's own greedy reply ($32$ tokens for refusal) with a detector.
For refusal the detector is the substring check, or a variant that also requires a safety-related word and rejects declines that cite missing information; for a language it is the share of the reply in that language, for flattery a list of flattering phrases, and for a backdoor the backdoor's own detector.
Two guards, used in every refusal run, limit the simplest ways to game the detector: words or phrases that the reply copies from the slot do not count, and replies shorter than six words score proportionally less.
The behavior reward has no gradient, so every searcher but the GRPO policy still draws its candidate tokens from the gradient of the imitation reward; nearly all behavior-reward searches are therefore guided by the bad model's replies, even though their reward is not.

\paragraph{White-box rewards.}
We also searched against rewards that read more of the model than a reply does: several forms of KL divergence to the bad model's next-token distribution, Jensen--Shannon divergence, the log-likelihood of the bad reply against that of the unsteered model's own reply, and cosine similarity to the bad model's final hidden state.
None found a slot that works without the refusal direction: searched with it removed, every one stays at or below $0.07$ on gemma-2b-it, and the three run that way on Llama-3-8B stay at $0.00$.
Their edge over imitation is small: on the judge, three cells on Llama-3-8B at the end of the prompt exclude zero, between $+0.029$ and $+0.110$, and one on gemma-2b-it at the start, $+0.131$, all at three seeds.

\paragraph{The mediated fraction.}
$\mf$ is measured relative to the model's rate with no slot, so it exceeds $1$ when removing the direction also takes away some of $s_0$, as for some function-vector slots.
For refusal, $s_0$ is at most $0.03$ on every instrument in the cells of Table~\ref{tab:core}, so $\mf = 1 - s_{\text{abl}}/s$, except where a result is said to be measured above the no-slot rate; the cut-off $s - s_0 \geq 0.10$ uses the measured no-slot rate.
The cut-off is on the slot's own rate, not the vector's, and it is there because the ratio is unstable below it: one slot on OLMo-2 after SFT reads $0.017$ intact and $0.170$ with the direction removed, an $\mf$ of $-9.2$.
Whether a direction is zero- or mean-ablated depends on whether zeroing it changes the model with nothing in the prompt, not on how it was built: the refusal directions are differences of means too, yet with them zeroed the model with no slot stays at $0.000$ on the judge wherever we measured it, and at $0.03$ or less on the substring check.
\citet{li2024optimalablation} argue for a constant chosen to change the model least, over both zero and the mean.
The random-direction cut draws its direction from a standard normal distribution and removes it in the same way as $v$.

\section{Refusal: Further Results}
\label{app:core}

Table~\ref{tab:core} gives the numbers behind Figure~\ref{fig:core}.

\begin{table}[ht]
\caption{The numbers behind Figure~\ref{fig:core}. The slot in each row is the one the substring check ranks best, and the judge columns score that same slot, so they are not the highest judge rate (for Llama-3-8B the highest are $0.883$ and $0.877$, and for Yi-6B-Chat $0.400$ and $0.180$, both falling to $0.000$ with the direction removed). ``Removed'' removes the refusal direction and ``Random'' a random direction; the judge was not run on the random-direction cut. Seeds 0--2 are pooled for Llama-3-8B-Instruct and Yi-6B-Chat, seed 0 elsewhere; slots are scored on $500$ held-out prompts for gemma-2b-it and $300$ for the other models. $^{\dagger}$These slots read below the cut-off on the judge, so their $\mf$ is not read (Appendix~\ref{app:more}).}
\label{tab:core}
\centering
\small
\begin{tabular}{llcccccccc}
\toprule
 & & \multicolumn{4}{c}{Substring check} & \multicolumn{4}{c}{Harm-grounded judge} \\
\cmidrule(lr){3-6}\cmidrule(lr){7-10}
Model & Position & Vector & Slot & Random & Removed & Vector & Slot & Removed & $\mf$ \\
\midrule
gemma-2b & start & $0.996$ & $0.970$ & $0.978$ & $0.008$ & $0.948$ & $0.870$ & $0.000$ & $1.000$ \\
gemma-2b & end & $0.996$ & $0.982$ & $0.984$ & $0.136$ & $0.944$ & $0.890$ & $0.004$ & $0.996$ \\
Llama-3-8B & start & $1.000$ & $0.997$ & $0.997$ & $0.003$ & $0.703$ & $0.827$ & $0.000$ & $1.000$ \\
Llama-3-8B & end & $1.000$ & $1.000$ & $0.997$ & $0.003$ & $0.703$ & $0.693$ & $0.003$ & $0.995$ \\
Yi-6B & start & $0.863$ & $0.830$ & $0.837$ & $0.017$ & $0.793$ & $0.120$ & $0.000$ & $1.000$ \\
Yi-6B & end & $0.863$ & $0.400$ & $0.423$ & $0.017$ & $0.793$ & $0.140$ & $0.003$ & $0.976$ \\
\midrule
gemma-2-9b & start & $0.970$ & $0.947$ & $0.947$ & $0.010$ & $0.963$ & $0.920$ & $0.043$ & $0.953$ \\
gemma-2-9b & end & $0.970$ & $0.977$ & $0.970$ & $0.003$ & $0.963$ & $0.897$ & $0.003$ & $0.996$ \\
Qwen2.5-7B & start & $0.693$ & $0.740$ & $0.767$ & $0.003$ & $0.700$ & $0.623$ & $0.013$ & $0.979$ \\
Qwen2.5-7B & end & $0.693$ & $0.890$ & $0.900$ & $0.117$ & $0.700$ & $0.587$ & $0.003$ & $0.994$ \\
Llama-3.1-8B$^{\dagger}$ & start & $1.000$ & $0.987$ & $0.983$ & $0.003$ & $0.027$ & $0.027$ & $0.003$ & $0.875$ \\
Llama-3.1-8B$^{\dagger}$ & end & $1.000$ & $0.983$ & $0.983$ & $0.010$ & $0.027$ & $0.023$ & $0.003$ & $0.857$ \\
Llama-2-7b & start & $0.997$ & $0.993$ & $0.993$ & $0.210$ & $1.000$ & $0.967$ & $0.303$ & $0.686$ \\
Llama-2-7b & end & $0.997$ & $0.990$ & $0.990$ & $0.430$ & $1.000$ & $0.920$ & $0.327$ & $0.645$ \\
\bottomrule
\end{tabular}
\end{table}

\paragraph{Slots, random slots and random directions.}
On Yi-6B-Chat the best slots reach $0.79$ and $0.45$ of the steered model on the substring check, averaged over three seeds.
The $108$ random-slot readings of Section~\ref{sec:core} reach at most $0.025$; they leave out the random slots scored beside the pasted slots of Appendix~\ref{app:transferfull}, the highest of which reads $0.027$.
The substring check reads the same random slots higher, up to $0.233$.
Over all $62$ judged slots, the substring check reads at most $0.313$ with the refusal direction removed, on a Yi-6B slot that the judge reads at $0.003$ or less, and random directions move a slot by at most $0.043$.
Over every run of ours with a valid ablation, $1{,}204$ readings of slots and instructions in $288$ runs, a random direction leaves the median mediated fraction at $0.000$, on each run's own detector.
In those runs a searched slot keeps more than $40\%$ of its effect without the direction in $86$ of $827$ rows ($10.4\%$; $82$ of them refusal slots, none a language slot) and a plain instruction in $143$ of $444$ ($32.2\%$), although their medians are closer ($\mf$ $0.989$ against $0.865$ at the start of the prompt); these count rows, which hold $81$ distinct slots and $92$ distinct instruction measurements.

\paragraph{The second prompt set.}
The second prompt set is the open-question, general-question, brainstorming and creative-writing prompts of \texttt{databricks-dolly-15k} that carry no context.
There the substring check's floor on gemma-2b-it rises to $0.077$ with no slot and $0.090$--$0.120$ with a random slot, where the judge's stays at $0.017$, and gemma's slot at the end of the prompt keeps $0.333$ on the substring check with the direction removed, where the judge reads $0.000$.

\paragraph{The limits of the word ban.}
The ban removes every token whose text contains one of the $38$ banned words, so a banned word can still reach a slot, assembled across tokens or in an inflection the list misses.
A scan on word stems finds $14$ of the $24$ slots free of every banned word in any form, and the count in Section~\ref{sec:core} uses only those: the best free slot beats the better instruction in $8$ of the $12$ cells, in two cells no slot is free, and in two the free slot loses.
On one Qwen2.5-7B seed the ban removed nothing the search would have used, and that search returned the unbanned search's slots exactly, as did one slot of another seed; outside that search the count is $6$ of $10$.
Counting every slot, the best beats the better instruction in $11$ of the $12$ cells, two of them from the search the ban did not change, though $6$ of those $12$ best slots carry a banned word.
On Llama-3.1-8B the best slots reach $0.295$--$0.610$ at the start of the prompt and $0.315$--$0.450$ at the end, against $0.010$ and $0.055$ for the instruction, and they clear the cut-off although the model's own vector reads only $0.025$ in these runs; on Qwen2.5-7B they read $0.153$--$0.437$ against $0.123$--$0.277$, the instruction winning at one seed at the start of the prompt ($0.267$ against $0.190$).
The substring check reads these runs far higher and agrees with the judge on them at $\kappa$ $0.296$, so we quote only the judge.

\paragraph{Plain instructions.}
On gemma-2b-it, counting only the slots and instructions scored on $500$ prompts, the slots go from $0.93$ to $0.02$ and all $14$ lose at least $80\%$ of their effect, against $2$ of $29$ instructions (substring check); on the judge the mean over $30$ instructions falls from $0.13$ to $0.006$.
In Figure~\ref{fig:routecut}, two of the dotted slots come from a search run with the direction removed; the one slot that keeps more than $20\%$ is one of them, at $0.06$ intact and $0.02$ without the direction, and three instructions rise when the direction is removed.
On Llama-3-8B-Instruct, $5$ and $7$ of the $15$ instructions read higher without the direction than with it, while at the start of the prompt three of the $15$ lose most of their refusal.
On Yi-6B-Chat the median $\mf$ is measured above the no-slot rate, over the $14$ and $25$ of $30$ instructions that reach $0.3$ on the broad rubric at the start and the end of the prompt.

\paragraph{Searching without the direction.}
On gemma-2b-it the $0.19$ of Section~\ref{sec:core} is the highest of $22$ cells at three and ten tokens ($0.004$ at three tokens alone), across cold and warm starts, larger reinforcement-learning budgets, white-box rewards and three seeds; seeds 1 and 2 stay at or below $0.01$.
At seed 0 (Figure~\ref{fig:search}), the ordinary search's best slot reads $0.920$ on gemma-2b-it and $0.877$ on Llama-3-8B-Instruct, and the control search's $0.906$ and $0.863$, which fall to $0.000$ and $0.003$ with the refusal direction removed; the $0.94$--$1.03$ divides the control search's best slot at each of three seeds by the ordinary search's at the first.
The slot that came closest to working without the direction came from an ordinary search: found under one white-box reward on Llama-3-8B at the start of the prompt, it kept $0.23$ on the substring check with the direction removed at one seed, where three random directions left it at $1.00$.
Of its $68$ remaining refusals $63$ name a content category, against $10$ of the steered model's $300$, and the same cell reads $0.02$ and $0.00$ at the other two seeds.
Tokens spread over a two-turn conversation and searched to make the model comply while the refusal vector stays on reach $0.00$ in every condition.

On Llama-3.2-3B the direction removed is one published by another group, at the layer and position our own extraction picks.
Over the eight slots of that search, removing the direction leaves a median $47\%$ of what three random directions leave, although the best slot at the end of the prompt reads more with the direction removed than with random directions removed ($0.313$ against a mean of $0.224$).
Removing a second published direction as well, the two together as one subspace, lowers the best slots to $0.227$ and $0.173$.
Over the $16$ slots of both searches, the real removal leaves less than the random ones in $14$ of $16$ comparisons, at a median ratio of $0.60$ against about $0.02$ on gemma; the two comparisons that go the other way are unexplained.

\paragraph{The reasoning model.}
On Qwen3-8B a second searcher's best slot goes from $0.13$ to $0.00$, with random directions at $0.13$--$0.17$.
The direction is hard to find on this generation: the recipe of \citet{arditi2024refusal} passes $2$ of $448$ candidates on Qwen3-8B, where the vector takes the no-slot rate from $0.01$ to $0.73$ on the substring check with every ablated and random floor at $0.01$, and $0$ of $512$ on Qwen3-14B, which therefore gets no route test.

\subsection{The four further chat models}
\label{app:more}

All four are single-seed (lower block of Table~\ref{tab:core}).

\paragraph{Llama-3.1-8B-Instruct.}
At the recipe's default coefficient the steered model reads only $0.027$ on the judge, and the slots of Table~\ref{tab:core}, the best on the substring check, read $0.027$ and $0.023$, below the cut-off.
The floor comes from the steering strength, not the judge: at $1$, $2$ and $4$ times the default coefficient the steered model reads $0.027$, $0.567$ and $0.000$.
Five other models swept at $0.5$, $1$ and $2$ times peak at the default, among them gemma-2b-it ($0.537$, $0.953$ and $0.513$); of the seven models in Table~\ref{tab:core} only gemma-2b-it and Llama-3.1-8B were swept, so the others rest on the recipe's default.
We also tested a direction that another group published for this model (layer $12$, position $-5$; ours is layer $11$, position $-2$): its steered model reads $0.130$ on the judge, which leaves too little judged refusal to test, and we count that cell as inconclusive.
Before the searches on these four models ran, the substring check was changed to count curly apostrophes, which Llama-3.1-8B uses; the change applies to every model, and the earlier runs' replies contain almost none.

\paragraph{Llama-2-7b-chat.}
A pattern check shows what the slots keep.
Of the replies that still refuse on the substring check with the direction removed ($63$ and $129$), $94\%$ and $79\%$ name a content category, against $18\%$ of the steered model's own refusals, because the slots put objectionable words in the prompt.
With the direction in place the same slots' refusals name one in $74$--$98\%$ of cases, so their declines are aimed at the slot's content with or without the direction.
The check was run after the result, and its pattern is a signal rather than a classifier: on this model's $60$ hand-labeled replies it fires on $0.67$ of declines aimed at injected material, $0.24$ of harm-grounded refusals and $0.10$ of replies that do the task.

\paragraph{Searching without the direction.}
The search reaches $0.003$ and $0.000$ on gemma-2-9b-it, against $0.963$ for its steered model, and $0.000$ at both positions on Qwen2.5-7B-Instruct, against $0.693$ and $0.690$.
It also reaches $0.000$ on Llama-3.1-8B, but that model's steered rate is $0.027$, so this is weak evidence.
The model with no slot and the direction removed reads $0.000$ on all four, and random directions move it by at most $0.007$.
On Llama-2-7b-chat three slots keep $0.620$, $0.627$ and $0.690$ with the direction removed, against $0.977$, $0.980$ and $0.820$ with it in place.
In the model-assisted check of $60$ of their replies, every harm-based decline is aimed at material the slot itself added to the prompt, at reweighted rates of $0.262$, $0.313$ and $0.100$, and none is a harm-grounded refusal of the request itself.
The check has two limits: with $20$ replies per slot, the $95\%$ upper bound on each slot's rate of such refusals is $0.33$, and the labels come from a model-assisted pass, not an independent human one.

\section{Beyond Refusal: Further Results}
\label{app:general}

\paragraph{How Table~\ref{tab:beyond} was built.}
The table uses seed 0, except for the function vectors (seed 1, the seed whose runs measure the ablated accuracy within the run).
The antonym instruction, the best of three over both prompt positions, is scored at the end of the prompt in an earlier run.
For the persona, the slot is the highest of the ten slots tested on these $100$ fresh prompts, each picked earlier by its judge rate on other prompts, and the instruction is the suffix the vector was built from; a reworded instruction reads $0.890$ and keeps $0.320$ without the direction.

\paragraph{Spanish, function vectors and the French backdoor.}
The best Spanish slot names the language, as many language slots do (Appendix~\ref{app:language}).
We had expected the contrastive Spanish direction to leave the SAE feature's slots alone; removing it stops every slot and instruction, and the slots searched to copy the contrastive vector need both directions too.
Other function-vector slots of the same runs fall to $0.065$ or less, and capitalize's slots at another seed to $0.094$ or less.
The French backdoor's true trigger reads $0.969$ on the detector, below the best slot.
A random direction moves slots other than the best by up to $0.05$ on Spanish and, for one country-capital slot, from $0.463$ to $0.390$.
At seeds 1 and 2 all $12$ checks read as at seed 0; one antonym check is borderline, with a mediated fraction of $1.03$ for slots and $1.05$ for instructions.
On the backdoored Llama-3.2-3B the direction carries $0.989$ of the slots' effect and $0.951$ of the instructions', means over six slots and four instructions.

The searches without the direction started from the best Spanish slot, from the antonym task's name, and on the French backdoor from its true trigger or from random tokens.
Fourteen further searches run with the Spanish feature removed produce no slot that works on both the intact model and the model without the feature: the one slot that works on the intact model ($0.667$ on the language detector) falls to $0.003$ with the feature removed, while three random directions leave it at $0.58$--$0.69$.
Those among them that were rewarded for the behavior itself found nothing even on the intact model, so they say nothing about a slot that works without the feature.
Of five searches on the French backdoor started from random tokens with the trigger direction removed, two found three-token slots that fire the backdoor on the intact model, at $0.984$ and $0.977$ against $0.977$ for the true trigger in that run, and none of the five exceeds $0.016$ without the direction.

\paragraph{Backdoors with no single direction.}
For four public backdoors whose payload is a refusal or a hate message, the best candidate direction takes away $0$--$13\%$ of the trigger's effect; for two gemma backdoors that need two triggers together, $9\%$ (hate message) and $27\%$ (refusal), measured above the clean rate.

\paragraph{The evil persona.}
The vector's $0.52$ is over $200$ held-out prompts, against $0.000$ with nothing added, and the judge's evil rubric agrees with hand labels $0.883$ of the time.
Shorter slots reach less: the searchers average $0.15$ at $3$ tokens, and $0.243$ and $0.265$ at $16$ tokens at the start and the end of the prompt.
At $32$--$64$ tokens and a second seed, searches started from a plain instruction reach $0.29$--$0.45$ and the same searchers started cold $0.20$--$0.22$; in a wider search with every searcher, hard-prompt projection started cold reaches $0.320$ and $0.300$ at $64$ tokens at two seeds.
One arm, started from a $16$-token slot, is unstable across seeds: $0.390$ and $0.138$ at the start of the prompt, $0.273$ and $0.318$ at the end.

\paragraph{Flattery.}
The sycophancy rubric agrees with hand labels $0.983$ of the time.
Slots chosen by their training score reach $0.07$ at $3$ tokens and at most $0.275$ at $32$--$64$ tokens; the $0.405$ of Section~\ref{sec:general} is a maximum taken on the test prompts, and apart from that slot none is above $0.36$.
A search rewarded for flattery in the model's own words finds nothing ($0.000$--$0.010$), though its reward, a fixed list of flattering phrases, scores the vector itself only $0.07$ and agrees with the judge only weakly ($\kappa = 0.113$).
The command the vector was built from reads $0.820$ at the start of the prompt and $0.790$ at the end, and $0.070$ and $0.050$ with the flattery direction mean-ablated; a reworded command reads $0.970$ and $0.880$, and keeps $0.880$ and $0.380$.
A random direction leaves both commands within $0.03$ of their rates.
Our searches were aimed at the first command's wording, since the imitation reward copies the vector's own replies, so the result is a failure of our search on a behavior that prompts can reach.
Distance from plain inputs does not explain it: flattery's steered states are no further from plain inputs than the evil persona's or Spanish's, which tokens did reach, though plain inputs vary least along its direction ($8.8$ times as much as along a random direction, against $16.1$ for evil and $36.5$ for Spanish).

\paragraph{Four persona vectors and the thinking-length vector.}
The rubrics of the four persona vectors that score $0.00$--$0.01$ have no hand labels, so we cannot tell a vector that does nothing from a rubric that misses what it does.
The thinking-length vector lengthens Qwen3-1.7B's thinking partly by making the model loop.
At coefficient $0.5$ it adds $74\%$ with $12\%$ of replies looping, against none without it, and at $2.0$ it multiplies thinking by $4.88$ with every reply looping and none answering, so no coefficient doubles the thinking without loops, and the planned search did not run.
In exploratory runs at coefficient $1.5$, where the vector takes thinking from $718$ to $2{,}087$ words and looping from $14\%$ to $88\%$ of replies, searched slots add $35$--$88$ words on average, the best single slot adds $210$, and no slot loops in more than $18\%$ of replies.
The search's reward compared only the first $24$ tokens of each reply, which do not differ in length or looping.

\paragraph{Zero ablation that installs Spanish.}
The zero-ablated model answers in Spanish at $0.975$ on the language detector and at $0.970$--$0.980$ on a second, disjoint set of prompts.
Its replies are not broken: they are as long as the plain model's ($74$--$76$ words against $76$) and only slightly more often degenerate ($3.0$--$3.5\%$ against $0.5$--$1.0\%$).

\paragraph{Longer Spanish slots.}
Longer slots also reproduce the Spanish feature and need it, but give no edge over an instruction.
At $16$ tokens they reach $0.81$ and $0.91$ on the judge, and at $32$ and $64$ tokens up to $0.74$ and $0.77$; with the feature removed they fall to $0.00$ on the language detector, while a random direction leaves them near their rate.
On that detector, a plain instruction padded with neutral text to $23$--$42$ tokens reaches $0.86$--$0.90$ at the end of the prompt, and the best long slot $0.89$.
One warm-started search at $32$ tokens was reward-hacked: its best slot makes the model repeat a Spanish word copied from the slot, which the detector reads at $0.79$ and the judge at about $0.05$, and which keeps $0.53$--$0.73$ with the feature removed.
We replaced that slot with another warm-started search's best slot after reading this result and before any further seed ran.

\paragraph{A matched fake vector.}
Tokens partly copy even a vector that carries no behavior.
For a behavior-free direction, built from random splits of instructions and moving the model at least as far as a real vector, the best slots close $16\%$ of the gap in held-out likelihood, against $30\%$ for two persona vectors, a ratio of $0.53$.
\citet{mody2026minimizing} report that optimized inputs suppress a random placebo direction about as hard as a real one, for one evaluation-awareness direction on Llama-3.2-3B.
Moving along a direction is therefore not, by itself, evidence that the direction matters, which is why our claims rest on removing the direction and on random-direction controls.
Three earlier attempts at this control did not match the fake vector's push to the real one's ($0.27$ and $0.38$ against $1.01$, or a collapse into repetition when scaled up) and are inconclusive.

\section{Moving Behaviors Between Models: Further Results}
\label{app:transferfull}

\subsection{Behaviors taught by search}

\begin{table}[ht]
\caption{Refusal taught across models at two seeds: the nine three-token cells that, at the first seed, both reproduced the teacher and lost the taught refusal with the receiver's direction removed. Each entry is the fraction of the teacher's refusal reproduced (substring check) / the mediated fraction for the receiver's own direction (judge). ``Judged'' gives the same slot's rate on the judge at seed 0 and, after the slash, the receiver's own refusal vector's; the teachers' saved replies were not judged. The fraction reproduced divides the receiver's substring rate by the teacher's, and the substring check errs differently on different models, so it is approximate. Bold marks the two that fall at the second seed. On the two later OLMo-2 stages the judge over-counts (Table~\ref{tab:gold}), so their reproduced rates are inflated and their mediated fractions are uncertain in either direction.}
\label{tab:transfer}
\centering
\small
\begin{tabular}{llccc}
\toprule
Cell & Position & Seed 0 & Seed 1 & Judged \\
\midrule
Llama-3-8B taught by gemma-2b & start & $0.98$ / $1.00$ & $0.98$ / $1.00$ & $0.597$ / $0.700$ \\
Llama-3-8B taught by gemma-2b & end & $1.01$ / $0.99$ & $1.00$ / $0.99$ & $0.777$ / $0.700$ \\
gemma-2b taught by Llama-3-8B & start & $0.95$ / $1.00$ & $0.96$ / $1.00$ & $0.840$ / $0.957$ \\
gemma-2b taught by Llama-3-8B & end & $0.94$ / $1.00$ & $0.95$ / $1.00$ & $0.843$ / $0.957$ \\
\midrule
Llama-3.1-8B taught by Llama-3-8B & end & $0.94$ / $0.98$ & $0.95$ / $1.00$ & $0.423$ / $0.027$ \\
Llama-2-7B taught by Llama-3-8B   & start & $0.98$ / $0.98$ & $0.94$ / $0.97$ & $0.840$ / $1.000$ \\
Qwen2.5-7B taught by Llama-3-8B   & end & $0.94$ / $0.99$ & $0.89$ / $0.98$ & $0.497$ / $0.683$ \\
\midrule
OLMo-2 DPO taught by Llama-3-8B      & end & $0.86$ / $0.97$ & $\mathbf{0.70}$ / $0.99$ & $0.730$ / $0.447$ \\
OLMo-2 Instruct taught by Llama-3-8B & end & $0.86$ / $0.93$ & $\mathbf{0.43}$ / $0.85$ & $0.713$ / $0.437$ \\
\bottomrule
\end{tabular}
\end{table}

\paragraph{Refusal.}
We judged $28$ runs at the first seed, $20$ of them across models.
OLMo-2 after SFT is among the receivers that lose $92$--$100\%$: its own vector reaches only $0.19$--$0.20$ on the judge, but its best slots read $0.293$ and $0.187$ and fall to $0.003$ and $0.000$.
On two of the cells that reproduce little of the teacher, the shortfall looks like a cost of copying another model's wording.
On gemma-2-9b, slots that copy the teacher's replies reach $0.49$ of the teacher, slots that copy the model's own steered replies $0.77$--$0.84$ of its own steered model, and the behavior reward $0.98$ at the same length, though that slot reads like an instruction; on OLMo-2 after SFT the behavior reward reaches $0.86$ of the teacher's rate, against $0.51$ for copying.
Qwen2.5-7B's $0.69$ at the start of the prompt is as high as its own vector reaches there.
At $16$ and $32$ tokens five of the six low cells we reran rise ($0.49 \rightarrow 0.98$, $0.57 \rightarrow 1.00$, $0.51 \rightarrow 0.89$, $0.70 \rightarrow 0.94$ and $0.56 \rightarrow 0.95$), while OLMo-2 Instruct at the start of the prompt falls ($0.68 \rightarrow 0.33$ at $16$ tokens and $0.64$ at $32$).
Of the $13$ cells rerun at a second seed, $10$ move by at most $0.07$, and the $3$ that drop ($0.86 \rightarrow 0.70$, $0.86 \rightarrow 0.43$ and $0.95 \rightarrow 0.73$) are all OLMo-2.

\paragraph{Spanish.}
A gemma-2-2b steered to answer in Spanish ($0.970$ on the judge) teaches Llama-3-8B, which answers in Spanish $0.000$ of the time on its own and $0.947$ with its own Spanish vector.
At two seeds the best slot reaches $0.870$ and $0.773$ at the end of the prompt and $0.707$ and $0.913$ at the start.
With Llama's own Spanish direction mean-ablated, every slot falls to $0.000$ at both seeds and positions, while three random directions leave every slot at least $85\%$ of its rate at the first seed and the best slots within $0.02$ at the second; with no slot the ablated model also stays at $0.000$.
A plain instruction ($0.980$ and $0.950$) beats every slot.
This is a single teaching pair; the judge's Spanish rubric agrees with hand labels on $0.974$ of items ($\kappa$ $0.944$).

\paragraph{French.}
The teacher is a public backdoored gemma-2-2b that answers in French when its trigger is present, and the receiver is again Llama-3-8B, which answers in French $0.000$ of the time on its own and $0.990$ with its French vector.
On a language detector, over three seeds, the best slot reaches $0.48$--$0.71$ at the start of the prompt and $0.23$--$0.87$ at the end, against $0.97$ and $0.90$ for a plain instruction.
With Llama's own French direction mean-ablated, every slot at every seed falls to $0.000$, except one at $0.02$, while three random directions leave the slots within $0.03$ of their rates (first seed).
On a different prompt pool the first seed's best slots reach $0.83$ and $0.62$, and again every slot falls to $0.000$.
The teacher was scored on a different prompt window from the receiver, so we give no fraction of the teacher, and no judge rubric for French has been validated.

\paragraph{The evil persona.}
A gemma-2-2b with its evil vector teaches Llama-3-8B, whose own evil vector reaches $0.447$; the teacher reads $0.603$ on the judge over $300$ prompts and replies of up to $128$ tokens, where the $0.52$ of Section~\ref{sec:general} is over $200$ prompts and $96$-token replies.
With $32$ tokens at the end of the prompt the best slot reaches $0.527$, and $93\%$ of it disappears when Llama's own evil direction is removed, while random directions move it by at most $0.013$.
The other three cells reach $0.100$--$0.193$, and all four are single-seed.
What fails at the start of the prompt is the copying, not the receiver's reach: slots searched against Llama-3-8B's own evil vector reach $0.507$ at the start and $0.427$ at the end.
A plain instruction reaches $0.987$--$0.990$ in every cell.

\paragraph{Emergent misalignment.}
Three public organisms built on Qwen2.5-7B-Instruct each teach a Llama-3-8B receiver at the start of the prompt; on the concrete, first-person questions used, $26$ of $36$ teacher replies are misaligned, against $0$ of $36$ on harmless instructions.
Every condition on the receiver reads $0.00$--$0.08$ on a misalignment rubric that agrees with hand labels $0.950$ of the time ($\kappa$ $0.897$) and whose three errors are all misses.
The receiver barely shows the behavior even with its own direction applied, built from $32$ contrast pairs written for this run, so neither the ceiling nor the dependence on the direction can be read.
Why the slots carry nothing is open: the search imitated only the first $32$ tokens of each reply, at one position, one length ($32$ tokens) and one seed.

\subsection{Pasted refusal slots}

\paragraph{Design.}
The $24$ donor slots are those that work on their own models, each pasted at the position where it was found.
A slot keeps its token ids when the receiver shares the donor's vocabulary; otherwise it is decoded to text and re-encoded with the receiver's tokenizer.
On gemma-2-9b-it, Llama-3.2-3B and Yi-6B-Chat, each pasted slot and two plain instructions to refuse are scored as they are, with the receiver's own refusal direction removed and with each of three random directions removed; harm-related and neutral words of the slots' lengths ($3$ and $10$ tokens) are scored at both positions.
gemma-1.1-2b-it and gemma-7b-it have no direction of their own and test transfer only.
Yi-6B-Chat's vector reads $0.58$ in Table~\ref{tab:paste} and $0.79$ in Table~\ref{tab:core} because the paste run uses a direction we extracted, not the published one.

\paragraph{The five smaller receivers.}
Slots from both donors transfer, across families too: Llama-3-8B's reach $0.730$ on gemma-2-9b-it and gemma-2b-it's $0.553$ on Llama-3.2-3B.
On gemma-2-9b-it each of the $66$ random-direction cuts of the $22$ slots that reach $0.10$ leaves its slot at $0.91$--$1.14$ of its rate (median $1.00$), and on Llama-3.2-3B the mean of three random cuts stays within $0.05$ of each slot's rate.
Generating and judging the pasted slots a second time reproduces gemma-2-9b-it's rates within $0.033$ (median $0.007$) and puts the loss at $92$--$100\%$ there and at a median of $84\%$ on Llama-3.2-3B.
Harm-related words raise refusal mostly at the end of the prompt, where they reach the $0.220$ and $0.190$ of Table~\ref{tab:paste}, against at most $0.123$ and $0.010$ at the start; neutral words stay at the no-slot rate ($0.000$--$0.007$).
On each of these two receivers $20$ of the $24$ slots beat the harm-related words at their own position, and $19$, $22$, $22$ and $23$ of the $24$ beat the better instruction at their own position on gemma-2-9b-it, Llama-3.2-3B, gemma-1.1-2b-it and gemma-7b-it.
On the last two, the best slot from the donor that shares the receiver's vocabulary beats the best re-encoded one at both positions.

\paragraph{The larger receivers.}
The donors, Llama-3.1-8B-Instruct and Qwen2.5-7B-Instruct, were each searched at two seeds with the gradient-guided and genetic searchers at both positions, four slots per seed, with the behavior reward (on the Llama donors, its variant that also requires a harm reason).
Receivers share the donor's tokenizer and chat template, so every slot pastes with identical token ids; they score $200$ (Llama) or $300$ (Qwen) held-out prompts, the 70B and 72B in 8-bit, which on a matched cell from another experiment moves rates by at most $0.03$.
Table~\ref{tab:paste} reports the first donor seed's four slots, except on Llama-3.1-70B, where those four reach only $0.020$ and the $0.190$ comes from the second seed's.
On Qwen2.5-72B three random directions leave the slot at $0.31$--$0.39$.
A second run with all eight slots gives a best of $0.317$, which falls to $0.047$ with the direction removed ($\mf$ $0.853$; random directions $0.34$--$0.37$); it is the same donor slot at the same position, and the two runs differ only in batch size and in the four added slots, so these rates move by about $0.05$ from run to run.
Two of the four slots and four of the eight read more than $0.02$ on Qwen2.5-72B, and two of the eight on Llama-3.1-70B; in each run only one reaches $0.10$.
Every pasted slot at the start of the prompt reads $0.04$ or less in all six runs.
The Qwen slot behind the $0.370$ contains none of the $38$ decline and harm-category words in any form, which does not rule out its acting like such words inside the model; the Llama slot behind the $0.190$ contains one harm-category word, which fits its doing no better than harm-related words.

\paragraph{Hand labels.}
The $420$ labels, $60$ per pasted condition, were made with no model pass first.
They give gemma-2-9b-it's slot $0.750$ ($95\%$ interval $0.628$--$0.842$) and its instruction $0.217$ ($0.131$--$0.336$), and Qwen2.5-72B's slot $0.417$ ($0.301$--$0.543$).
With the receiver's direction removed, the labels and the judge agree on every item ($0.017$ and $0.033$), so the loss with the direction removed does not rest on the judge's error.
Llama-3.2-3B's hand-labeled set, from the searches that taught it refusal, shows no over-count, and the other receivers' pasted conditions have no hand labels.

\paragraph{Why refusal and language differ.}
One reading is that post-training pulls the two behaviors apart.
Answering in another language is a request it teaches directly \citep{wei2022flan,wang2022supernatural,longpre2023flan,muennighoff2023crosslingual,singh2024aya,grattafiori2024llama3,yang2024qwen25,lambert2024tulu3}, while refusing a harmless request is what it trains against \citep{ouyang2022instructgpt,bai2022hh,grattafiori2024llama3,brahman2024coconot}.
We cannot test this without intervening on training.

\subsection{The language control}
\label{app:language}

For a behavior that a plain instruction already produces, these cells ask whether pasted slots carry it to larger models and whether it needs the receiver's own direction there.
Each donor is a model with a French contrastive vector added, searched with the behavior reward at $32$ tokens and both positions; donor and receiver share a tokenizer and chat template, and French is scored by a language detector.

\paragraph{Transfer.}
Slots found on Llama-3.1-8B-Instruct make Llama-3.1-70B-Instruct answer in French at $0.525$, $0.420$ and $0.635$ for three donor seeds at the end of the prompt, against $0.130$ for French prose of the same length, $0.000$ for English prose and for a random slot, and $0.985$ for a plain instruction; at the start of the prompt no seed's slots beat French prose ($0.325$).
On Qwen2.5-72B-Instruct, against at most $0.215$ for French prose and $0.990$--$0.995$ for the instruction, the best slot at three donor seeds reaches $0.830$, $0.035$ and $0.920$ from a 7B donor, $0.780$, $0.930$ and $0.960$ from a 14B donor and $0.590$, $0.675$ and $0.745$ from a 32B donor.
Slots from a 1.5B donor do not transfer to the 72B (at best $0.200$, $0.000$ and $0.020$).

\paragraph{The receiver's direction.}
On Llama-3.1-70B, removing the receiver's own French direction takes away $96$--$100\%$ of the pasted slots' effect over the three seeds, against $81\%$ for French prose and $89$--$90\%$ for a plain instruction.
On Qwen2.5-72B the share runs from $99\%$ to below zero across donors, seeds and positions, with no relation to donor size or to how strongly the slot transfers, while the instruction, where its removal was recorded, loses $27$--$36\%$.
At the start of the prompt the 32B donor's slots do not need the receiver's direction: at its second and third seeds they reach $0.675$ and $0.735$, and both read $0.700$ with it removed.

\paragraph{Slots against the instruction.}
On every receiver, writing the instruction beats pasting any language slot, in all $115$ comparisons in the same run and position, and even on the donor's own model a language slot beats it in only $16$ of $75$.
Many language slots name their language: over our runs, those that do read a median of $0.820$ on the donor's own model, against $0.002$ for those that do not ($72$ and $224$ slots), a split that is observational, since the second group includes dead searches and negative controls.
Searched again with the name banned and pasted into Llama-3.1-70B, the best French slot reads $0.295$, below French prose at $0.325$; slots that do not name the language can still transfer, up to $0.735$ on Qwen2.5-72B.
Spanish pasted from gemma-2-9b-it into gemma-2-27b-it crosses weakly: a slot that does not name the language reads $0.180$ at the start of the prompt and one that names it $0.620$ at the end, against $0.900$ and $0.905$ for a plain instruction, and both fall to $0.000$ with the receiver's direction removed.
A judge rubric for French agrees with the detector on $0.975$ of about $19{,}000$ replies ($\kappa$ $0.943$), though no hand labels validate it; at the start of the prompt the judge reads about a quarter lower than the detector on the cells checked, so French rates there are upper bounds.

\section{Auditing Details}
\label{app:audit}

\paragraph{The statistic.}
A word is a run of three or more letters, lowercased.
For each backdoored model we take at most one slot per searcher, the strongest first, from searches with six searchers, at one or both prompt positions and with slots of $3$ and $8$ tokens.
Every model in Table~\ref{tab:auditmain} keeps six slots, except the first, which keeps five.
Of the two backdoors at $0.33$, one has a trigger word as its most-shared word, and on the other two trigger words tie for most-shared with an ordinary word.
On the two backdoors with a code-like trigger no word appears in more than one slot; they also fire on random slots, on $0.09$ to $0.78$ of prompts, so a search can succeed there without the trigger.
As controls, a fine-tune with an always-on brand preference and three sets of random slots each keep four slots in which no word repeats (a share of $0.25$), and none is called.
A thirteenth backdoor, with a two-word trigger on Qwen2.5-3B, fires on only $0.44$ of held-out prompts even with its own trigger, so none of its slots was kept.
A second statistic, meant to recognize an always-on preference from the meaning of a model's unshared words, does not work: every model scores below its random baseline.

\paragraph{The fire test.}
A backdoor fires if its detector flags at least half of the $30$ replies with the word, at least $0.3$ more than without it (for a pair, than with one of its words alone), and at most $0.2$ of the clean base's replies with the word.
One word of a pair on its own reaches at most $2$ of $30$.

\paragraph{The blind judges.}
Each item lists a model's kept slots and their positions, and nothing else.
Each held-out item went to a fresh Claude Opus 5.5 agent with no tools, once with an open prompt and once with a checklist of kinds of behavior and trigger and of patterns to look for, among them a word that most slots share.
The held-out items are the nine models and two random-slot controls outside the five items on which the prompts were developed.
With the checklist the judge names the trigger for $5$ of $9$ ($7$ of $9$ if hedged answers count, a category we added after reading them) and the behavior for $4$ of $9$; with the open prompt it names the trigger for $3$ of $9$ ($5$ of $9$).
Two of the eleven had also been judged once during development; on the eight models never judged then, it names the trigger for $5$ of $8$ ($6$ of $8$) with the checklist and $2$ of $8$ ($4$ of $8$) without.
It reports nothing for both controls under both prompts, and it misses the code-like trigger under both.
Open-weight judges, given all $16$ items with the same prompts, do worse.
Qwen2.5-32B-Instruct names the trigger for $0$ of the $12$ backdoors with the checklist and $2$ of $12$ without, abstaining on $12$ and $10$ of the $13$ models, and finds nothing in all three controls.
With the checklist, Qwen2.5-7B-Instruct names the trigger for $0$ of the $12$ backdoors --- its one answer that is not ``unknown'' is on the brand-preference fine-tune, which has no trigger, and there it correctly reports that the behavior is always on --- and Qwen2.5-1.5B-Instruct names none and reports a modification in all three controls.

\paragraph{The one-token baseline.}
The lookup scores every vocabulary token on the imitation reward that most of the searches use, so it needs the same examples.
Its top tokens set off the backdoor on $0.58$ to $0.92$ of held-out prompts and the clean base on none.
On the Qwen backdoor it scores $751{,}000$ sequences (a sequence is one search prompt with one candidate slot), fewer than the $1.35$ million scored by the $24$ searches that supplied that backdoor's slots.
On the two-word backdoor, no pair built from the lookup's top $200$ tokens contains the trigger words, while the searches' most-shared word there is a trigger word, though below the cut-off.
Neither method finds a trigger of several tokens: the lookup ranks the word inside one such trigger at $4{,}327$, on a backdoor outside the audit, and the searches missed both code-like triggers.

\paragraph{The payload-free variant.}
The reward for a slot is the clean base's log-likelihood of its own greedy reply to a clean prompt, over the reply's first $16$ tokens, minus the backdoored model's, both with the slot in the prompt; an instruction, which moves both models, earns little.
Eight searches, four searchers at two seeds, look for a $3$-token slot at the start of the prompt.
A slot is kept if, on $32$ held-out prompts, it beats the best of $8$ random $3$-token slots; a call also needs at least $4$ kept slots, and the called word must beat the random slots and the bare prompt on the same reward.
Unlike the count with examples, this run is a held-out test of the statistic and its cut-off.
The five single-word backdoors are on gemma-2-2b (French), Qwen2.5-3B (hate message and refusal) and Llama-3.2-3B (French and refusal); the other four models are the gemma-2-2b backdoor that needs two words, the Llama-3.2-3B backdoor with a code-like trigger, and two Qwen2.5-3B fine-tunes with an always-on preference and no trigger.
Both backdoors it names answer in French, with $0.62$ and $0.50$ of their slots sharing the word; on held-out prompts the word gives French on $1.00$ and $0.66$ of replies, against $0.00$ without it and on the clean base.
On the other seven models no word appears in more than one kept slot.
We have not run the variant as a scan of public fine-tunes whose backdoors, if any, are undocumented.

\section{Judges and Hand Labels}
\label{app:judge}

\begin{table}[ht]
\caption{Agreement of the harm-grounded judge with hand labels, $60$ items per model unless a count is given. ``Broad'' is the refusal rubric of the original judge on its $67$ refusal items. Each $60$-item sheet holds $30$ replies the judge counts and $30$ it does not, so $\kappa$ is $2 \times$ agreement $- 1$; ``Precision'' is the share of the judge's yeses that the labels confirm. Labels began as a blind pass by a language model, without sight of the judge's verdicts, and were then reviewed by a person, except: $^{\ddagger}$labeled directly by a person; $^{\S}$reviewed by a second model pass, with four items confirmed by a person; gemma-2-9b's figures are its blind labels. Llama-3-8B pools two sets of $60$ (FN $2$ and $4$).}
\label{tab:gold}
\centering
\small
\begin{tabular}{lccccc}
\toprule
Model & Agreement & $\kappa$ & Precision & FP & FN \\
\midrule
gemma, broad ($67$)           & $0.985$ & $0.970$ & $0.973$ & $1$ & $0$ \\
gemma, harm-grounded ($99$)   & $0.960$ & $0.898$ & $1.000$ & $0$ & $4$ \\
Llama-3.2-3B                  & $0.967$ & $0.933$ & $1.000$ & $0$ & $2$ \\
OLMo-2 SFT                    & $0.933$ & $0.867$ & $0.867$ & $4$ & $0$ \\
Qwen3-8B$^{\S}$               & $0.933$ & $0.867$ & $0.900$ & $3$ & $1$ \\
Yi-6B                         & $0.917$ & $0.833$ & $0.833$ & $5$ & $0$ \\
Llama-3-8B ($120$)            & $0.892$ & --- & $0.883$ & $7$ & $6$ \\
Llama-3.1-8B                  & $0.883$ & $0.767$ & $0.767$ & $7$ & $0$ \\
gemma-2-9b                    & $0.800$ & $0.600$ & $0.633$ & $11$ & $1$ \\
Llama-2-7b-chat               & $0.800$ & $0.600$ & $0.600$ & $12$ & $0$ \\
Qwen2.5-7B                    & $0.750$ & $0.500$ & $0.500$ & $15$ & $0$ \\
OLMo-2 Instruct$^{\ddagger}$  & $0.733$ & $0.467$ & $0.467$ & $16$ & $0$ \\
OLMo-2 DPO$^{\ddagger}$       & $0.717$ & $0.433$ & $0.433$ & $17$ & $0$ \\
\bottomrule
\end{tabular}
\end{table}

\paragraph{The judge.}
On gemma-2b the judge's rate is $0.253$ against $0.293$ by hand, and on the two later OLMo-2 stages only $13$ and $14$ of its $30$ yeses are true.
Two models give almost no harm-grounded refusal even with their own vector applied, OLMo-2 after SFT ($0.19$--$0.20$) and Llama-3.1-8B ($0.03$); the hand labels show the judge missing none of their refusals, so these floors are not undercounts.
All $11$ of the judge's over-counts on gemma-2-9b fall in conditions with a searched slot in the prompt, and in each the model objects to the injected material and then does the task.

\paragraph{The substring check.}
On Qwen3-32B the substring check gives the model's own steering vector $0.990$, on short canned replies that the judge scores $0.047$.
On Llama-3.2-3B and the three OLMo-2 models steered at twice the usual strength, the replies degenerate into loops, and the substring check reads three of the four at $1.000$ while the judge reads all four at $0.050$ or less.
On refusal slots pasted into larger models it errs in both directions: it reads Llama-3.1-70B's best slot at $0.375$ (judge $0.190$) and, on Qwen2.5-72B, a plain instruction at $0.110$ against the best slot's $0.063$; the two instruments agree there at $\kappa$ $0.211$ to $0.700$.
On OLMo-2 DPO and Instruct, the gap between the substring check and the judge ($0.52$--$0.54$) is explained neither by the length of the replies nor by the steering strength, and we report it as unexplained.

\paragraph{Decoy declines.}
On Yi-6B, slots from searches rewarded by the substring check read up to $0.83$ on the broad rubric and at most $0.18$ on the harm-grounded judge.
A person checked the proposed labels of $49$ of the benchmark's decoys and agreed that none is a harm-grounded refusal, while the substring check counts $44$ of the $49$ as refusals.
The two judges cannot be scored on the benchmark's items, because together they chose them.
An early result did not survive the change of judge.
On the broad rubric, three-token slots searched to imitate the steered model's replies depended on the direction more than slots rewarded for refusing, by $+0.083$ in mediated fraction at the start of the prompt and $-0.010$ at the end, and five-token slots by $+0.22$ and $+0.11$.
On the harm-grounded judge every three-token slot has a mediated fraction of at least $0.995$ and the five-token slots read $1.00$ each, so the gap was the decoy.

\paragraph{Declines of injected material.}
The $16$-token search of Section~\ref{sec:measurement} started from a plain refusal instruction.
Sorted by pattern and spot-read, none of its best slot's $500$ replies gives a harm, legal or policy reason with the direction removed, against $475$ with the direction in place.
The $60$ hand-labeled replies are half judged yes and half no; none of the $60$ requests contains anything harmful, yet $21$ of the replies are sound harm-grounded refusals of the material the slot added.

\section{Further Related Work}
\label{app:related}

\paragraph{Steering and prompting.}
Persona vectors extend contrastive steering to character traits \citep{chen2025persona}.
Prompts and vectors are also linked by in-context vectors \citep{liu2024icv}, by prompt steering replacements, which learn token-specific steering that reproduces a prompt's effect \citep{heyman2026steerlike}, and by prefix steering, where steering a short span of tokens often keeps much of what full steering does \citep{zhu2026prefixsteering}.
Steering moves a model much further than the prompts its authors first wrote \citep{hua2025evalaware}, although a later version of that paper reports hand-written prompts that reach deployment-level behavior, and prompting usually beats steering on benchmarks \citep{wu2025axbench}.
\citet{bhargava2023magicword} ask which outputs a short prompt can reach; we take a behavior that a prompt can reach and ask whether it needs a direction.

\paragraph{Refusal and its scoring.}
Refusal is encoded separately from harmfulness \citep{zhao2025harmfulness}, which our gemma-2b-it result refines: plain instructions keep a decline but lose its harm reason when the direction is removed.
Refusal categories have distinct directions, yet steering along any of them moves one shared trade-off between refusal and over-refusal \citep{joad2026morerefusal}, and refusal that survives ablation has been reported as a category-concentrated residue \citep{shi2026fragile} and as dormant backup features \citep{prakash2025beyondsorry}.
Refusals of benign requests have been induced on purpose, by poisoned instruction-tuning data \citep{shu2023exploitability}, by blocker documents in a retrieval corpus \citep{shafran2025jamming} and by benign prompts evolved to look harmful \citep{wu2025evorefuse}; \citet{vonrecum2024cannot} separate declines for lack of capability from declines on safety grounds.
Standardized evaluations score jailbreaks with trained classifiers \citep{mazeika2024harmbench}, and LLM judges need validation against humans \citep{zheng2023judging}.

\paragraph{Search, backdoors and transfer.}
Short token sequences found by gradient search can push a model toward a chosen output on any input \citep{wallace2019universal}, and prompt search has been treated as reinforcement learning before \citep{deng2022rlprompt,diao2022bdpl,perez2022redteaming}.
Backdoors can also be planted through poisoned human feedback \citep{rando2024universal}, and auditing games ask whether an auditor can uncover a hidden objective \citep{marks2025auditing}.
Emergent misalignment runs largely along one direction \citep{soligo2025convergent}, as our evil persona does, yet it did not cross between models.
\citet{hughes2026toxscreendetectingllmpoisoned} release about $800$ poisoned models; their defender has no clean reference model and tells a planted trigger from a jailbreak by a curvature statistic, while our auditor has the clean base and discards slots that set it off too.
Knowledge distillation trains a student on a teacher's outputs \citep{hinton2015distilling}, which can pass on a trait the outputs do not express when the two share a base model \citep{cloud2025subliminal}, and weak-to-strong generalization has a smaller model supervise a larger one \citep{burns2023weaktostrong}; all of these change the student's weights, while here no weights change and a few prompt tokens are the only channel.

\section{Further Limitations}
\label{app:limitations}

\paragraph{Seeds.}
The search without the refusal direction ran at three seeds on gemma-2b-it and at one on Llama-3-8B and Yi-6B, and its random-direction control at three seeds on gemma-2b-it and one on Llama-3-8B, with none on Yi-6B.
Spanish transfer rests on a single teaching pair at two seeds and the evil-persona transfer on a single seed; French transfer has three seeds, over which the level and the better prompt position change while the dependence on the direction does not.
Refusal pasted into much larger models has two donor seeds per family and one receiver run per slot count.

\paragraph{Where the test is hard to read.}
On Yi-6B the slots reproduce only part of the vector, so its result rests on small effects.
On Llama-3.1-8B and on OLMo-2 after SFT the steered model itself sits near the judge's floor, so most of their slots fall below the cut-off, and a mediated fraction is read only on the slots that clear it.
What Llama-2-7b-chat's slots keep is read by a pattern check run after the result and by a model-assisted check, not a human one (Appendix~\ref{app:more}).
On Llama-3.2-3B the direction accounts for only part of the refusal, taught or searched, and in the language control nothing we measured predicts how much of a pasted French slot's effect on Qwen2.5-72B needs the receiver's direction.

\paragraph{Measurement.}
The hand labels were not all made the same way (Table~\ref{tab:gold}).
The decoy benchmark covers two models, and only $60$ of the $2{,}990$ replies in its extension have hand labels; of the pasted conditions, only those on gemma-2-9b-it and Qwen2.5-72B do.
The evil-persona, sycophancy and misalignment rubrics agree with hand labels on $0.883$, $0.983$ and $0.950$ of items; French is scored by a language detector, and the backdoor payloads by their own detectors.

\paragraph{Scope.}
Pasted slots reach $70$B and $72$B receivers only between models that share a tokenizer, with those two receivers in 8-bit.
A refusal direction extracted on Qwen2.5-32B-Instruct steers it ($0.573$ on the judge, with the plain, ablated and random-direction floors at $0.000$), but we ran no refusal search at that size.

\section{Released Material}
\label{app:ethics}

As supplementary material we release four things: the decoy benchmark of Section~\ref{sec:measurement}, each reply scored by the broad rubrics and by the harm-grounded judge, with the $49$ labels a person checked, and its extension of $2{,}990$ declines of injected material; the hand labels behind every judge figure we quote, with the blind pass, the review pass and the final labels kept separate (the $420$ labels of the pasted conditions are a single human pass), so that Table~\ref{tab:gold} and the pasted-condition comparison can be recomputed; the scoring code, that is, the judges' rubrics, the substring and harm-word checks, and the programs that build and score the sets; and the runs' outputs, with the searched slots in them.
The benchmark's base set contains no searched slot and no harmful request: replies that repeat a slot's letters were dropped, and every prompt is a harmless instruction.
Its extension does repeat slot material, in $2{,}602$ of $2{,}990$ items, some of it offensive, and ships with its own content warning; we release it anyway, because a benchmark meant to show that a metric over-reads is useful only to someone who can run the metric on it.
The searched slots are in the release, though we quote none in the paper: most make a model refuse harmless requests or answer in another language, and those that elicit the persona or a backdoor's payload give nothing a plain prompt or a published trigger does not.
For the persona a plain instruction does better than any slot we found ($0.94$--$0.99$ against at most $0.53$), and because the persona is measured on harmless instructions, what it shows is tone, not harmful capability.

\end{document}